%% file: main.tex
\documentclass{article}
\usepackage[linesnumbered, ruled,vlined]{algorithm2e}
\usepackage{arxiv}
\usepackage[utf8]{inputenc} % allow utf-8 input
\usepackage[T1]{fontenc}    % use 8-bit T1 fonts
\usepackage{hyperref}       % hyperlinks
\usepackage{url}            % simple URL typesetting
\usepackage{booktabs}       % professional-quality tables
\usepackage{amsfonts}       % blackboard math symbols
\usepackage{nicefrac}       % compact symbols for 1/2, etc.
\usepackage{microtype}      % microtypography
\usepackage{lipsum}		% Can be removed after putting your text content
\usepackage{graphicx}
\usepackage{amsmath}
\usepackage{nicefrac} % compact symbols for 1/2, etc.
\usepackage{microtype} % microtypography
\usepackage{xcolor} % colors
\usepackage{algpseudocode}
\usepackage{multirow}
\usepackage{caption}
\usepackage{subcaption}
\usepackage{wrapfig}
\usepackage{amsthm}
\usepackage{color}
\usepackage{mathtools}

\DeclareMathOperator*{\argmax}{argmax}

\newtheoremstyle{mytheoremstyle} % name
 {\topsep} % Space above
 {\topsep} % Space below
 {\itshape} % Body font
 {} % Indent amount
 {\bfseries} % Theorem head font
 {.} % Punctuation after theorem head
 {.5em} % Space after theorem head
 {} % Theorem head spec (can be left empty, meaning ‘normal’)
\newtheoremstyle{mydefinitionstyle} % name
 {\topsep} % Space above
 {\topsep} % Space below
 {} % Body font
 {} % Indent amount
 {\bfseries} % Theorem head font
 {.} % Punctuation after theorem head
 {.5em} % Space after theorem head
 {} % Theorem head spec (can be left empty, meaning ‘normal’)

\theoremstyle{mytheoremstyle}

\theoremstyle{mytheoremstyle}
\newtheorem{proposition}{Proposition}
\theoremstyle{mytheoremstyle}

\theoremstyle{mydefinitionstyle}
\newtheorem{definition}{Definition}
\newcommand{\titlefont}{\fontsize{16}{20}\selectfont}

\usepackage{Definitions}

\newcommand*\samethanks[1][\value{footnote}]{\footnotemark[#1]}

\begin{document}

\title{Training Free Graph Neural Networks for Graph Matching}

\author{Zhiyuan Liu% \thanks{Institute of Data Science, National University of Sinagpore}
\thanks{School of Computing, National University of Sinagpore} \\ acharkq@gmai.com
 \And Yixin Cao\thanks{School of Computing, Singapore Management University} \\ caoyixin2011@gmail.com\\
 \And Fuli Feng\samethanks[1] \\ fulifeng93@gmail.com \\
 \And Xiang Wang\samethanks[1] \\ xiangwang@u.nus.edu \\
 \And Jie Tang\thanks{Computer Science Department, Tsinghua University} \\ jietang@tsinghua.edu.cn\\
 \And Kenji Kawaguchi\samethanks[1] \\ kenji@comp.nus.edu.sg\\
 \And Tat-Seng Chua\samethanks[1] \\chuats@comp.nus.edu.sg}

% \authornote{Institute of Data Science, National University of Sinagpore}
% \authornote{School of Computing, National University of Sinagpore}
% \email{acharkq@gmai.com}
% \affiliation{\institution{}\country{}}

\date{}
\maketitle

\begin{abstract}
    We present a framework of \textbf{T}raining \textbf{F}ree \textbf{G}raph \textbf{M}atching (\textbf{TFGM}) to boost the performance of Graph Neural Networks (GNNs) based graph matching, providing a fast promising solution without training (training-free). TFGM provides four widely applicable principles for designing training-free GNNs and is generalizable to supervised, semi-supervised, and unsupervised graph matching. The keys are to handcraft the matching priors, which used to be learned by training, into GNN's architecture and discard the components inessential under the training-free setting. Further analysis shows that TFGM is a linear relaxation to the quadratic assignment formulation of graph matching and generalizes TFGM to a broad set of GNNs. Extensive experiments show that \textit{GNNs with TFGM} achieve comparable (if not better) performances to their fully trained counterparts, and demonstrate TFGM's superiority in the unsupervised setting. Our code is available at \href{https://github.com/acharkq/Training-Free-Graph-Matching}{https://github.com/acharkq/Training-Free-Graph-Matching}.
\end{abstract}

\input{./introduction}

\input{./related_works}

\input{./methodology}

\input{./experiment}
\section{Conclusion and Future Work}
% conclusion
We present the first framework TFGM for graph matching with training-free GNNs. TFGM provides a fast approximate solution to graph matching without training. We handcraft matching priors, which used be learned by training, into GNN's architecture and show that TFGM is a linear relaxation to graph matching's QAP formulation. Extensive experiments demonstrate TFGM's generalizability to supervised, semi-supervised, and unsupervised graph matching. Further ablation studies validate TFGM's key components. 
% future work
In the future, we are interested in exploring training-free GNNs for zero-shot learning.

\bibliography{reference}
\bibliographystyle{plainnat}

\clearpage
\appendix
\input{./appendix}

\end{document}

%% file: introduction.tex
\section{Introduction}
% TFGM sidesteps two crucial problems when training GNNs: 1) the limited supervision due to expensive annotation, and 2) training's computational cost. 
Graph matching aims to find equivalent nodes between graphs while respecting the compatibility of node features and graph structures~\cite{wang2019learning}. It is crucial in many real-world applications, such as to match keypoints on images~\cite{zanfir2018deep}, find equivalent entities between knowledge graphs (KGs)~\cite{sun2017cross}, and link users across different social networks~\cite{zhang2019multilevel}. 
% Remarkably, graph matching also faces large-scale data due to the usage in search engines~\cite{yoon2021image,ling2021deep}.
% Graph Neural Networks (GNNs) have shown promising performance on learning latent representations of nodes and edges, revolutionizing the techniques for graph matching.  
% Various Graph Neural Network (GNN) based methods have achieved great success for graph matching due to their ability to abstract node features while preserving structures~\cite{zanfir2018deep, yan2020learning, fey2020deep, wu2020neighborhood}. Nevertheless, it is non-trivial to train a GNN. This is because, 
To better abstract node features for matching, it has become a \textit{de facto} standard to train GNNs in supervised or semi-supervised models~\cite{fey2020deep,sun2020benchmarking}. There are two key difficulties in the training process. On one hand, graph matching's annotation is often limited~\cite{berrendorf2021active} and sometimes unavailable~\cite{saraph2014magna}. This is because graph matching's annotation is generally labor-intensive due to the large candidate space~\cite{zhong2018colink}. In specific domains, annotation is further barriered by issues like cross-language (\textit{e.g}, KGs~\cite{chen2016multilingual}), incomplete profile (\textit{e.g.}, social networks~\cite{zhong2018colink}) or lacking expertise knowledge (\textit{e.g.}, protein networks~\cite{ficklin2011gene}).
On the other hand, training GNNs is computationally expensive due to the exponentially growing number of neighbors with depth~\cite{chiang2019cluster, zou2019layer}. 

\begin{table}[t] % {r}{60mm}
\centering
   \small
    \caption{Results on graph matching benchmark DBP{\small ZH-EN}~\cite{sun2017cross}. GNNs share the same architecture (except residual connection), but differ in the weights.}
    \begin{tabular}{lc}
    \toprule
    \textbf{Model} & \textbf{Accuracy (\%)} \\
    \midrule
    NodeMatch & 60.3 \\
    Training-free GNN w/o residual & 37.2 \\
    Training-free GNN w residual & 62.5 \\
    Fully trained GNN & 70.4 \\
    \bottomrule
    \end{tabular}%
    \label{tab:preliminary}%
    \vspace{-0.5cm}
\end{table}

These difficulties accompanied with training raise the following \textbf{R}esearch \textbf{Q}uestions. \textbf{RQ1}: Is it possible to conduct graph matching using GNNs without training? \textbf{RQ2}: To what extent, can we apply such training-free setting in real applications? \textbf{RQ3}: What are the differences when designing training-free neural architectures?
% Can the training-free setting benefit from the existing advances in neural architecture? 
To resolve these questions, we present the \textbf{T}raning \textbf{F}ree \textbf{G}raph \textbf{M}atching (\textbf{TFGM}) framework for graph matching with GNNs without training. We propose and justify four principles to design training-free GNNs that are generalizable to supervised, semi-supervised, and unsupervised graph matchings. 

\textbf{RQ1}.
We show that TFGM is a linear assignment problem (LAP) relaxation of graph matching's quadratic assignment problem (QAP) formulation~\cite{caetano2009learning, cho2013learning}, which can preserve both node compatibility and structural compatibility. 
The preliminary results in Table~\ref{tab:preliminary} demonstrate our analysis: a training-free GNN outperforms NodeMatch, which directly compares node features without any structural information. As expected, training-free GNNs perform worse than the fully trained GNN, posing a great challenge of incorporating knowledge that trained models can learn into the training-free framework. 

\textbf{RQ2}. We generalize the unsupervised TFGM to supervised, semi-supervised settings by incorporating annotations without training. Annotation contains crucial knowledge of matching. We let training-free GNNs use annotation to match fully trained GNNs' performance. The general principle is to leverage annotation to generate more discriminative node representation: for the supervised graph matching, we design a label-discriminative feature to mine potential alignment signals from annotation data via kNN search; for the semi-supervised graph matching, we propose a node feature initialization strategy to incorporate alignment signal.

\textbf{RQ3}. We propose two rules of training-free neural architectures, inspiring four design \textbf{P}rinciples for GNNs. 1) Handcraft matching priors, which used to be learned by training, into GNN architecture. The first rule explains that powerful architecture leads to better performance (\textbf{P4}). It also implies using annotation (\textbf{P2}). We further let node embedding to preserve neighbors of different localities (\textbf{P1}) to enable a strict matching of neighbors in the same order. 2) Discard the components inessential for training-free. We remove the nonlinearity between layers and propose weight-free GNN to eliminate the noise caused by random weights (\textbf{P3}).

For evaluation, we extensively experiment on three benchmarks, including (supervised) Keypoint Matching~\cite{zanfir2018deep}, (semi-supervised) Entity Alignment~\cite{sun2017cross}, and (unsupervised) Protein-Protein Interaction Network Alignment~\cite{vijayan2015magna++}. We test TFGM with GraphSAGE~\cite{hamilton2017inductive}, SplineCNN~\cite{fey2018splinecnn}, and DGMC~\cite{fey2020deep}.
Experimental results support our analysis and demonstrate the effectiveness of TFGM.
TFGM achieves significant improvements than unsupervised models and performs comparable (if not better) to fully trained GNNs in supervised and semi-supervised settings. Ablation studies verify TFGM's main components and the efficient training-free property. 

% In addition, we show that TFGM is flexible to almost arbitrary GNNs, including SplineCNN, which has a nonlinear kernel function. 
%TFGM works as a bridge to transfer the many accumulated inductive bias in neural networks to areas where deep learning cannot attain.
% fixme: Our basic idea is that TFGM works as a bridge to transfer many accumulated inductive bias in neural networks to areas where deep learning cannot attain.

% Our contributions can be summarized as follows:
% \begin{itemize}
%     \item We propose a basic framework for graph matching with training-free GNNs backed by theoretical analysis;
%     \item We propose TFGM to preserve neighbors of different orders and leverage annotation in supervised and semi-supervised settings without training to enhance the basic framework;
%     \item We experiment TFGM with three popular GNNs for supervised, semi-supervised, and unsupervised graph matching to verify our analysis and demonstrate superior performance.
% \end{itemize}

%% file: related_works.tex
\vspace{-0.2cm}
\section{Related Work}
% In this section, we briefly review recent researches on graph matching and training-free neural networks.

\subsection{Graph Matching}
% We present the advances in graph matching and their applications. Note that graph matching has been separately investigated under different names in various practical settings. We use the name graph matching as much as possible for expression consistency.

\noindent\textbf{Technical Route}.
% The development of Graph Matching, from combinatorial solvers to deep learning models
Graph matching is mathematically formulated as an NP-hard QAP. Thus, initial works focus on relaxation to deal with the intractability~\citep{conte2004thirty}. In practice, graph matching can be eased by comparing graph attributes. 
To better measure the similarity of attributes, SVM, CNN, and GNN are successively introduced for graph matching~\cite{caetano2009learning, zanfir2018deep, wang2019learning, Xu2019CrosslingualKG}. Initially, deep learning methods are introduced to obtain better feature representations. A natural idea for improvement is thus to embed existing combinatorial solvers into neural networks~\cite{wang2019learning, sarlin2020superglue}. This idea has been applied on graph matching's LAP relaxation by applying the differentiable Sinkhorn networks~\cite{cuturi2013sinkhorn, mena2017sinkhorn}. ~\citet{rolinek2020deep} further combine GNNs with an advanced QAP solver based on Lagrange decomposition~\cite{swoboda2017study, swoboda2019convex} for graph matching. Meanwhile, GNNs are also studied as combinatorial solvers instead of feature extractors~\cite{wang2019neural, fey2020deep}. Our work shares similar spirits. We show that GNNs can be used to relax the QAP to make the problem tractable. More importantly, we show that GNN's fitness to graph matching is independent of training.
% fixme: Training-free GNN is firstly shown by Kipf and Welling~\cite{kipf2016semi}: random-weight GCN can generate node embeddings that preserve distances between nodes in the Karate club network~\cite{zachary1977information}. Veli{\v{c}}kovi{\'c} et al.~\cite{velivckovic2019deep} further show that the graph embedding obtained from random-weight GCN can be applied for node classification.

% \yixin{I'd prefer to survey related work according to unsupervised, semi-supervised and supervised. briefly explain their settings and applications. mainly talk about semisupervised including social network alignment and entity alignment.}
\noindent\textbf{Applications}.
% We briefly review graph matching's application in three representative practical problems: \textit{keypoint matching}, \textit{entity alignment} and \textit{social network alignment}.
Graph matching has various applications, including social networks, KGs, and Computer Vision (CV). Social networks are large graphs with rich topological patterns. Graph matching models rely on the \textit{isomorphic assumption} and aim to maximize the structural consistency~\cite{zhang2015multiple, zhang2019multilevel}. We can leverage attributes~\cite{yan2021bright} and additional networks~\cite{chu2019cross} for improvement. 
Similarly, Entity Alignment combines structures and attributes to find equivalent entities between KGs. Knowledge graph embedding models and relation-aware GNNs are adopted to learn the heterogeneous graphs~\cite{sun2020benchmarking, zhao2020experimental}. In CV, graph matching is applied to find the semantic equivalent keypoints between different objects~\cite{zanfir2018deep} and the same object's points but from different perspectives~\cite{sarlin2020superglue}. Geometric prior is studied to benefit object tracking~\cite{chen2001multi}, pose estimation~\cite{girdhar2018detect}, and point cloud registration~\cite{wang2019deep}. In these applications, training has been a long-standing issue due to the limited annotation~\cite{chen2016multilingual, zhou2021attent} and scalability~\cite{zhu2020collective,mao2021boosting}. Thus, we seek solutions from training-free strategies. 

\subsection{Training-free Neural Networks}
\noindent\textbf{Graph-Augmented MLPs}.
TFGM is related to Graph-Augmented MLPs (GA-MLPs)~\cite{ chen2019powerful, wu2019simplifying}. GA-MLPs obtain structure-aware node embeddings by applying a set of graph operators on node features. This step is training-free because graph operators are dependent only on structure, \textit{e.g.}, the adjacency matrix. Further, GA-MLPs train a classifier on top of structure-aware node embeddings for downstream tasks and have achieved very competitive performance to full GNNs. Because the training is independent of graph structure, GA-MLPs are trivially scalable to large graphs~\cite{frasca2020sign}. The idea is firstly introduced in SGC~\cite{wu2019simplifying}, which uses the power of a normalized adjacency matrix as the graph operator. The graph operator is demonstrated to be a low-pass filter~\cite{wu2019simplifying, nt2019revisiting}. \citet{frasca2020sign} enlarge the family of graph operators.~\citet{chen2020graph} theoretically show that the VC-dimension
% ~\cite{vapnik1971uniform} 
of existing GA-MLPs grows poly-exponentially with the number of layers. ~\citet{zambon2020graph} show that graph distance defined by training-free GNNs is metric. TFGM is different from previous works in that we focus on graph matching, which requires solving a combinatorial optimization problem.

% This phenomenon is explained by GCN's connection to the Weisfeiler-Lehman test of graph isomorphism~\cite{weisfeiler1968reduction}, which is also known as exact graph matching. This connection justifies the feasibility of graph matching with training-free GNN. In this work, we present GNN's relation to the inexact graph matching for attributed graphs. We also present a framework that can take advantage of training data.

\noindent\textbf{Neural Networks with Random Weights}.
TFGM is ideally similar to the Neural Networks with Random Weights (NNRW), which initiates from Random Vector Functional Link networks~\cite{igelnik1995stochastic}. It is proved that NNRW, in which weights between the input and hidden layers are randomly assigned, are universal approximators~\cite{li1997comments, huang2006universal}. The works on CNNs with random filters further expand NNRW to practical applications of object detection and image restoration~\cite{jarrett2009best, saxe2011random, ulyanov2018deep}. These works demonstrate that NNRW can be used to obtain meaningful representation that is linearly separable for downstream tasks. Inspired by previous works, we propose TFGM to obtain graph representation with training-free GNNs. 
% Previous works still require weight updating in at least the final layer for prediction.

%% file: methodology.tex
% \vspace{-0.2cm}
\section{Methodology}
\label{sec:method}
In this section, we introduce TFGM for the unsupervised graph matching, followed by TFGMws that generalizes TFGM to supervised and semi-supervised graph matching.
% \vspace{-0.2cm}
\subsection{Problem Formulation}
\label{sec:problem}
We define a graph as $\mathcal{G} = (\mathcal{V}, A, X, E)$, where $\mathcal{V}=\{1,2...\}$ is the set of nodes, $A\in \{0,1\}^{|\mathcal{V}|\times |\mathcal{V}|}$ is the adjacency matrix, $X\in \mathbb{R}^{|\mathcal{V}|\times d_x}$ is the node feature, and $E\in \mathbb{R}^{|\mathcal{V}|^2 \times d_{e}}$ is the edge feature. $d_x$ and $d_e$ are the dimension sizes for features of nodes and edges. Given graphs $\mathcal{G}^{(s)}=(\mathcal{V}^{(s)}, A^{(s)}, X^{(s)}, E^{(s)})$ and $\mathcal{G}^{(t)}=(\mathcal{V}^{(t)}, A^{(t)}, X^{(t)}, E^{(t)})$, \textit{w.l.o.g.}, let $|\mathcal{V}^{(s)}|\leq |\mathcal{V}^{(t)}|$, graph matching~\cite{caetano2009learning, cho2013learning} can be formulated as a QAP:
\begin{equation}\label{eq:gm}
 S^* = \argmax_{S \in \Tcal} \sum_{\substack{i\in \mathcal{V}^{(s)}\\ j\in \mathcal{V}^{(t)}}}Q_{ij}S_{ij} + \sum_{\substack{i,i'\in \mathcal{V}^{(s)}\\ j,j'\in \mathcal{V}^{(t)}}}T_{ii';jj'}S_{ij}S_{i'j'},
\end{equation}
where $S\in \{0,1\}^{|\mathcal{V}^{(s)}|\times |\mathcal{V}^{(t)}|}$ is an assignment matrix such that $S_{ij}=1$ iff $i\in \mathcal{V}^{(s)}$ is mapped to $j\in \mathcal{V}^{(t)}$. The entry $Q_{ij}$ of the matrix $Q\in \RR^{|\mathcal{V}^{(s)}|\times |\mathcal{V}^{(t)}|}$ measures nodes' similarity based on their features $X_i^{(s)}$ and $X_j^{(t)}$. $T_{ii';jj'}$ measures similarity between edge features $E_{(i,i')}^{(s)}$ and $E_{(j,j')}^{(t)}$.
% \fuli{My first impression on this notation is the i-th row, i'-th column of E }
The set $\Tcal=\{S\in \{0,1\}^{|\mathcal{V}^{(s)}|\times |\mathcal{V}^{(t)}|}:[\forall j\in \mathcal{V}^{(t)}, \sum_{i\in \mathcal{V}^{(s)}} S_{ij} \leq 1]\wedge[\forall i\in \mathcal{V}^{(s)}, \sum_{j\in \mathcal{V}^{(t)}} S_{ij} = 1]\}$ represents the assignment constraints to guarantee a one-to-one mapping. In Equation~(\ref{eq:gm}), the first linear term is to preserve the compatibility between nodes; the second quadratic term is to preserve the compatibility between edges.

If not noted, we apply the GCN-flavored definition of GNNs:
\begin{align}
 l=0 &:\enspace \enspace \enspace  \mathrm{GNN}_{l}(A, X)=H^{(l)}=X, \\
 l=1,\dots,L &:\enspace  \begin{cases}
    \mathrm{GNN}_{l}(A, X) = A H^{(l-1)}W_l, \\
    H^{(l)} = \sigma(\mathrm{GNN}_{l}(A, X)),
    \label{eq:gnn}
 \end{cases}
\end{align}
where $\sigma$ is a nonlinear activation function. $\{W_l\}_1^L$ is a series of trainable matrices. If the GNN is training-free, all matrices in $\{W_l\}_{1}^L$ are randomly sampled from $\frac{1}{\sqrt{d}}\mathcal{N}(\mathbf{0}, \mathbf{I})$. $d$ is the dimension of the hidden layer. This definition can be easily generalized to existing GNNs~\cite{kipf2016semi,hamilton2017inductive} by replacing the adjacency matrix $A$ with the corresponding graph operators.

\subsection{Training Free Graph Matching}
\label{sec:tfgm}

Like other machine learning methods, graph matching with GNNs has two phases: training and inference. In the training phase, the weights in GNNs are optimized to pull the embeddings of equivalent nodes together and push different nodes away. In the inference stage, we compute the similarities between node embeddings generated by the fully trained GNNs for graph matching. In light of the preliminary studies~\cite{chen2020graph} and experiments (Table~\ref{tab:preliminary}), we hypothesize that the weights in GNNs are not crucial for graph matching, but the majority comes from the inductive bias in GNN's architecture. We, therefore, propose the BasicTFGM that skips the training phase and directly applies randomly initialized GNNs for graph matching inference.

\begin{definition}[BasicTFGM] \label{def:1}
 Given an arbitrary training-free graph neural network $\phi: \mathcal{G} \mapsto \phi(\mathcal{G})\in \mathbb{R}^{|\mathcal{V}|\times d}$, \textit{BasicTFGM with $\phi$} is defined by finding the assignment $S^*$ that maximizes the dot-product of corresponding nodes' embeddings:
 \begin{equation}
 \label{eq:objective}
 S^* = \argmax_{S \in \Tcal} \sum_{i\in \mathcal{V}^{(s)}, j\in \mathcal{V}^{(t)}} S_{ij}\left(\phi(\Gcal^{(s)})\phi(\Gcal^{(t)})\T\right)_{ij}.
 \end{equation}
%  \vspace{-0.3cm}
\end{definition}
% The BasicTFGM measures node similarity generated embeddings 
From Table~\ref{tab:preliminary}, we observe that training free GNN's performance drops by $25\%$ if residual connection is disabled, showing the importance of lower-order information. The BasicTFGM only measures the similarities between nodes concerning their $L$-th order neighborhoods. The lower-order information may be crucial but ``washed out'' during feedforward. We, therefore, propose TFGM for improvement.

\begin{definition}[TFGM] \label{def:2}
 For an arbitrary $L$-layer graph neural network $\phi_L$, let $\phi_l (0\leq l\leq L)$ be the first $l$ layers of $\phi_L$. \textit{TFGM with $\phi_L$} is defined by finding the assignment matrix $S^*$ that maximizes the cosine similarity of corresponding node embeddings from all GNN layers as:
\begin{equation}\label{eq:cos_gm}
 S^* = \argmax_{S\in \Tcal} \sum_{\substack{i\in \mathcal{V}^{(s)}\\ j\in \mathcal{V}^{(t)}}}S_{ij}\left(\sum_{l=0}^L \mathrm{Cos}\left(\phi_l(\Gcal^{(s)}), \phi_{l}(\Gcal^{(t)})\right)\right)_{ij}.
\end{equation}
% \vspace{-0.3cm}
\end{definition}

TFGM enhances the BasicTFGM by preserving different orders of neighbor information and also the initial node feature ($l=0$) to facilitate a comprehensive similarity measurement. Further, because the norms of graph embeddings can change over different layers, we normalize the node embeddings at every layer by replacing dot-product with cosine similarity. TFGM guarantees the strict matching of neighbors from the same order and enforces the prior that information of different orders should weigh the same. The similarity measurements of different orders are summarized into a single linear term, guaranteeing to be solvable within polynomial time. Compared to the BasicTFGM, TFGM is more robust to over-smoothing because it considers the neighborhoods of all $0,1,...,L$ orders, wherein the over-smoothing issue usually happens in high-order layers that cannot affect the comparisons of low-order layers.

% \vspace{-0.2cm}
\subsection{Utilizing Annotation without Training}
\label{sec:no_training}
So far, TFGM can well suit the unsupervised graph matching. We now generalize TFGM to the supervised and semi-supervised settings. The annotation data can supervise graph matching models to perform better feature extraction~\cite{cho2013learning} and support the matching of uncertain nodes with close-by ground truth matchings~\cite{rocco2018neighbourhood}. To this end, we introduce two strategies for TFGM to utilize such priors without training. The enhanced TFGM is termed as TFGMws (TFGM with supervision).

The general principle of both strategies is to leverage annotation to generate more discriminative node representations. 
We present the core ideas of the strategies. Detailed algorithms are in Appendix~\ref{appendix:alg}. Given two graphs $\mathcal{G}^{(s)}$ and $\mathcal{G}^{(t)}$, both settings aim to find an assignment $S^*$ that corresponds to the equivalent mapping between $\mathcal{V}^{(s)}$ and $\mathcal{V}^{(t)}$. 

\noindent \textbf{Supervised Graph Matching} provides node labels for graphs in a training set $\mathcal{D}$~\cite{bourdev2009poselets, everingham2010pascal, cho2013learning}. Unlike node classification, the node labels are mutually exclusive within one graph. For graph matching, nodes with the same label but from different graphs are marked as equivalent. 

\begin{wrapfigure}[23]{rth}{0.5\textwidth}
    % \begin{figure}[t!]
            \centering
            \begin{subfigure}[b]{0.5\columnwidth}
                \centering
                \captionsetup{size=footnotesize, labelfont=footnotesize}
                \includegraphics[width=\columnwidth]{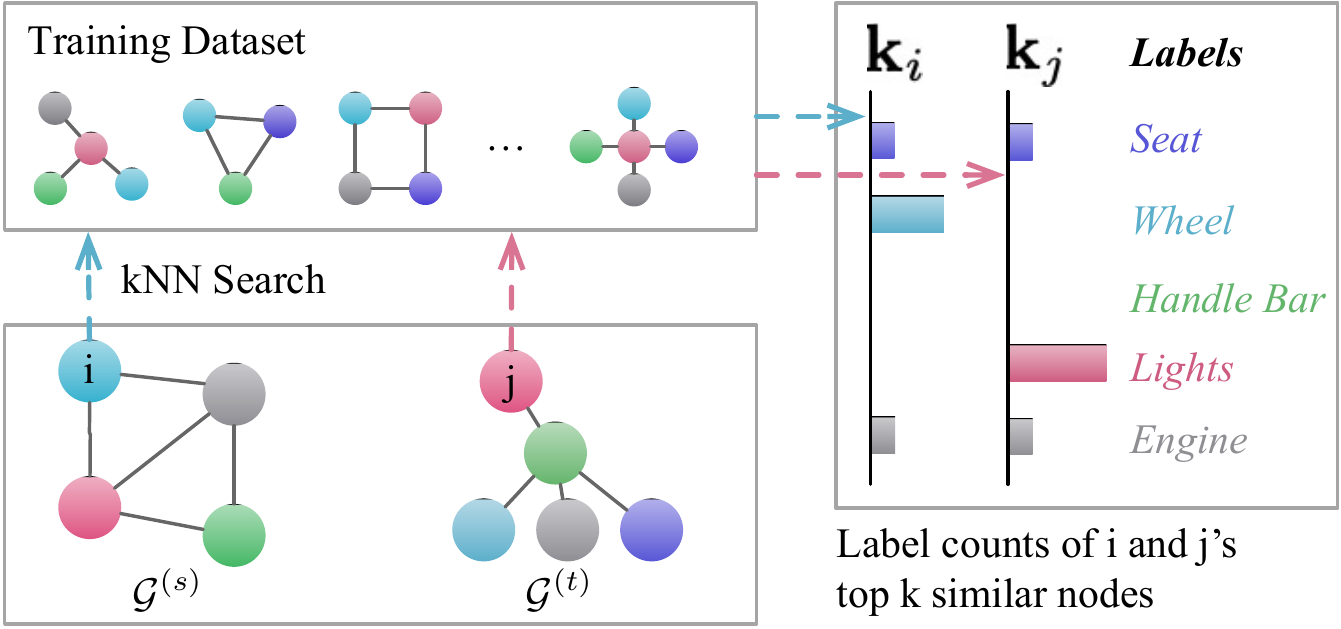}
                \caption{Supervised setting. Generate node i and j's label-discriminative feature $\mathbf{k}_i$ and $\mathbf{k}_j$ with kNN search. Different colors denote different node labels. Example labels are from a motorbike.}
                \vspace{0.3cm}    
            \label{fig:knn}
            \end{subfigure}
            \hfill
            \begin{subfigure}[b]{0.5\columnwidth}
              \centering
              \captionsetup{size=footnotesize, labelfont=footnotesize}
              \includegraphics[width=\columnwidth]{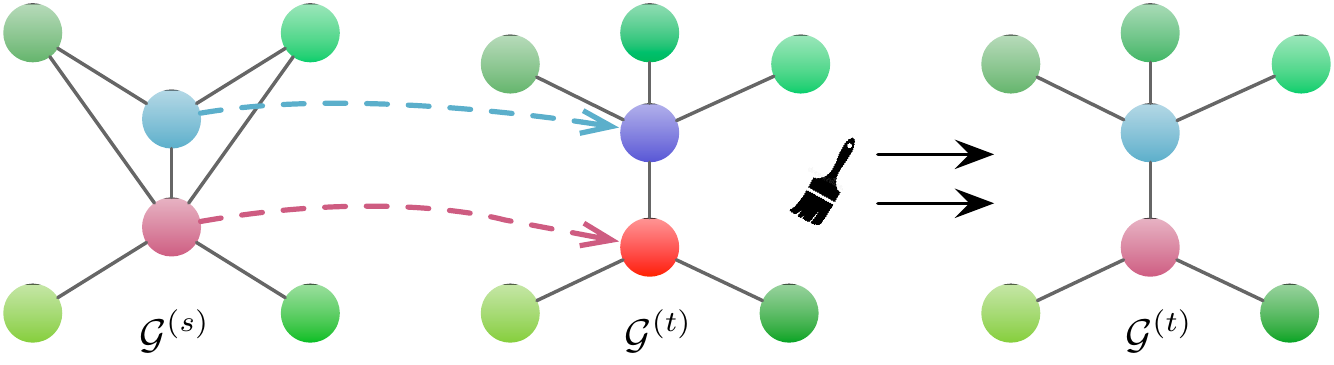}
            %   \vspace{0.01cm}
                \caption{Semi-supervised setting. Force known equivalent nodes (connected by dashed arrows) to have the same initial node features. }
            \label{fig:force_signal}
            \end{subfigure}
          \caption{Diagrams for utilizing annotation without training.}
        \label{fig:use_annotation}
    \end{wrapfigure}

To utilize annotation without training, we generate a label-discriminative feature $\mathbf{k}_i$ for all node $i$ in $\mathcal{G}^{(s)}$ and $\mathcal{G}^{(t)}$ by performing a kNN search in training dataset (Figure~\ref{fig:knn}). The basic idea is to find $i$'s $k$ closest nodes in training dataset

and use the labels of these $k$ nodes as $\mathbf{k}_i$. If $i$ is a \textit{wheel} of a motorbike, $\mathbf{k}_i$ will have a high \textit{wheel} score. $\mathbf{k}_i$ captures the priors in annotation and is more discriminative than the node's original feature. Thus, we solve the LAP with node similarity measured by $\mathrm{Cos}(\mathbf{k}_i, \mathbf{k}_j)$ ($\forall i\in \mathcal{V}^{(s)}, j\in \mathcal{V}^{(t)}$).

We present the steps to generate $\mathbf{k}_i$ for node $i$ in an arbitrary graph $\mathcal{G}$. The strategy is a variant of Matching Networks~\cite{vinyals2016matching}. 
1) We conduct graph matching between $\mathcal{G}$ and every graph in the training dataset $\mathcal{D}$ using the unsupervised TFGM, and thus obtain the similarity scores between $i$ and every node in the training dataset.
2) For each graph in the training dataset, we keep only the most similar node to $i$.
3) Select the top $k$ nodes $\mathcal{L} = \{v_1, v_2, ... , v_k\}$ that have the largest similarity scores to $i$.
4) Let $\mathbf{k}_i = \sum_{v\in \mathcal{L}} \mathbf{y}_v$, where $\mathbf{y}_v$ is the one-hot encoding of node $v$'s label.

\noindent \textbf{Semi-supervised Graph Matching} provides a set of annotated equivalent node pairs $\mathcal{I}=\{(i,j)| i\in \mathcal{V}^{(s)}, j\in \mathcal{V}^{(t)}\}$ as supervision. The goal is to find equivalent nodes among the unannotated nodes in $\mathcal{G}^{(s)}$ and $\mathcal{G}^{(t)}$. To incorporate annotation, we set the initial features of nodes in $\mathcal{I}$ to be the same (Figure~\ref{fig:force_signal}): for all $(i,j)\in \mathcal{I}$, let $X^{(s)}_i\rightarrow X^{(t)}_j$ or vice versa. The intuition is that GNNs can propagate the strong equivalence signal, \textit{i.e.}, $\mathrm{Cos}(X_i^{(s)}, X_j^{(t)})=1$, to neighbors of $i$ and $j$, and thus support their matching. This strategy operates on the feature initialization phase and adds $O(|\mathcal{I}|d)$ computation overhead, which is much cheaper than training.

\section{Design Principles and Justification}
\label{sec:justification}
\subsection{Design Principles of TFGM}
We summarize our framework as four design \textbf{P}rinciples to boost the performance of training-free GNNs for graph matching. They enable training-free GNNs to perform comparably to their fully trained counterparts.
\begin{itemize}
    \item \textbf{P1}: Concatenate the normalized embeddings from all GNN layers as node representations (Section~\ref{sec:tfgm});
    \item \textbf{P2}: If annotation is available, use it to generate more discriminative node representations (Section~\ref{sec:no_training});
    \item \textbf{P3}: For GCN-flavored GNNs, \textit{e.g.}, $\mathrm{GNN}_l(A,X)=AH^{(l-1)}W_l$, where $A$ is a graph operator. Use their weight-free versions, \textit{e.g.}, $\mathrm{GNN}_l(A,X)=AH^{(l-1)}$;
    \item \textbf{P4}: Explore more powerful graph operators/kernels.
\end{itemize}
\textbf{P1} and \textbf{P2} are intuitively explained in previous section. \textbf{P3} is to eliminate the random noise caused by random weights. \textbf{P4} encourages the exploration of more appropriate inductive bias for the dataset. Pioneer study~\cite{saxe2011random} and our experiments show that better architecture leads to better performance even in the training-free setting. 

In the rest of this section, we justify TFGM by showing that it is a linear relaxation of graph matching's QAP formulation. Further, we show that random-weight GNN is an unbiased estimator of weight-free GNN for graph matching if nonlinearity is removed (\textbf{P3}). Finally, we generalize TFGM to more GNNs (\textbf{P4}).
% We justify of \textbf{P3} and \textbf{P4} later in this section. 

\subsection{TFGM is a Linear Relaxation of the QAP}
\label{sec:tfgm_linear}
As stated in Equation~(\ref{eq:gm}), graph matching is a QAP, which is NP-hard. We now show that \textit{BasicTFGM with GNN} is a linear relaxation of the QAP.
Inspired by the neighborhood consensus~\cite{rocco2018neighbourhood, fey2020deep}, we relax the quadratic term in Equation~(\ref{eq:gm}) to linear as follows:
\begin{equation}
 \label{eq:new_gm}
 S^* = \argmax_{S \in \Tcal} \sum_{\substack{i\in \mathcal{V}^{(s)}\\ j\in \mathcal{V}^{(t)}}}Q_{ij}S_{ij} + \sum_{\substack{i,i'\in \mathcal{V}^{(s)}\\ j,j'\in \mathcal{V}^{(t)}}}A_{ii'}^{(s)}A_{jj'}^{(t)}P_{i'j'}^{(ij)}S_{ij},
\end{equation}
where $T_{ii';jj'}$is measured by $A^{(s)}_{ii'}A^{(t)}_{jj'}$, a binary indicator for the existence of edges $(i,i')$ and $(j,j')$; one of the assignment $S_{i'j'}$ is replaced by its approximate estimation $P_{i'j'}^{(ij)}\in \mathbb{R}^{(|\Vcal^{(s)}|\times |\Vcal^{(t)}|)^2}$, which is a 4-dimensional tensor. We give $P$ a superscript $(ij)$ (\textit{i.e.}, two extra dimensions) to maintain $S_{i'j'}$'s dependence on index $(ij)$ -- because $S_{i'j'}$ is an optimization variable, its value is dependent on the other optimization variable $S_{ij}$. In other words, this superscript allows $P$ to measure the similarity between $i'$ and $j'$ while considering the matching of other nodes. To summarize, Equation~(\ref{eq:new_gm}) holds a similar objective to Equation~(\ref{eq:gm}) -- find an assignment matrix $S^*$ that maximizes the node compatibility and structural compatibility.

% Note that, $P$ does not have to obey the matching constraints to be a good estimation. 
% While $Q_{ij}$ measures the node compatibility, the assignment approximation $P^{(ij)}_{i'j'}$ together with $A^{(s)}_{ii'}A^{(t)}_{jj'}$ measures the structural similarity between node $i$ and $j$. 
Note that, Equation~(\ref{eq:new_gm}) becomes an LAP, if $Q$ and $P$ have no weights to be optimized. Thus, we can solve it within polynomial time with a theoretical guarantee. Now, the core problem is to find suitable functions that can produce a good estimation of $Q$ and $P$. However, is it feasible to find good compatibility measurements $Q$ and $P$ that 1) require no training and 2) can measure equivalence between nodes and structures? The answer is yes. Let us first focus on the node compatibility measurement $Q$. When nodes have real-world features, such as text names and image patches, thanks to the development of pre-trained models, non-parametric functions (\textit{e.g.}, dot-product and cosine similarity) based on embeddings of the real-world features can largely measure the semantic equivalence~\cite{reimers2019sentence}. 

Next, we move forward to the structural compatibility $P$ by applying the \textit{BasicTFGM with GNN}.
\begin{proposition}
 \label{proposition:1}
 BasicTFGM with $\mathrm{GNN}_L(A, X)$ is equivalent to solve Equation~(\ref{eq:new_gm}) with $Q=0$ and $P=(H^{(L-1)})^{(s)}W_L ((H^{(L-1)})^{(t)}W_L)\T$.
\end{proposition}
The proof is in Appendix~\ref{sec:proofs}. Proposition~\ref{proposition:1} shows that \textit{BasicTFGM with GNN} is a special case of solving Equation~(\ref{eq:new_gm}): BasicTFGM sets node compatibility $Q=0$ and uses a lower dimensional $P$. This shows training-free GNN's potential to approximate the structural term in graph matching. 

Due to the BasicTFGM's limitation, we have proposed TFGM to preserve neighbor information of different orders and recover the node compatibility $Q$. For any matrix $M$, we define $M_i$ to be the transpose of its $i$-th row vector. Let $\|\cdot\|_2$ represent the Euclidean norm and denote by the symbol ``$\odot$'' the element-wise multiplication. We show that \textit{TFGM with GNN} is equivalent to solve Equation~(\ref{eq:new_gm}) with a specific choice of $P$ and $Q$. Thus, we can summarize that TFGM is a linear relaxation of the QAP.
\begin{proposition} \label{proposition:2}
 \textit{TFGM with $\mathrm{GNN}_L(A, X)$} is equivalent to solve Equation~(\ref{eq:new_gm}) with $Q=Z^{(0)} \odot (X^{(s)}(X^{(t)})\T)$ and $P^{(ij)}= \sum_{l=1}^L Z^{(l)}_{ij}(H^{(l-1)})^{(s)}W_l ((H^{(l-1)})^{(t)}W_l)\T $, where $Z^{(l)}\in \RR^{|\mathcal{V}^{(s)}|\times |\mathcal{V}^{(t)}|} $ is the normalization matrix. $Z^{(l)}_{ij}=1/(\|\mathrm{GNN}_{l}(A^{(s)}, X^{(s)})_{i}\|_{2} \|\mathrm{GNN}_{l}(A^{(t)}, X^{(t)})_{j} \|_{2})$ for $l=0,\dots,L$.
\end{proposition}

\subsection{Random-weight GNN \textit{v.s.} Weight-free GNN (P3)}
\label{sec:rand_vs_free}
GA-MLPs~\cite{chen2020graph} show that graph operators without nonlinear activation function can extract discriminative representations for node and graph classification. Based on their results, we conjecture that \textit{it is inessential to use the nonlinear activation in training-free GNNs for graph matching}. 
Besides, one of the most important reasons for using activation function is to let neural networks learn complex nonlinear patterns. This reason does not hold when weights are not trained. 
Thus, we remove GNN's nonlinear activation function to investigate linear graph operators' effectiveness for graph matching. Specifically, we define the following random-weight GNN and weight-free GNN:
\begin{alignat*}{2}
 & \text{Random-weight}\quad && \mathrm{GNN}_L(A, X) = A^L X W_1 ... W_L; \\
 & \text{Weight-free}\quad && \mathrm{GNN}_L(A,X) = A^L X.
\end{alignat*}
% This definition can be generalized to GCN and GraphSAGE by replacing the graph operator $A$ to $\tilde{D}^{-1/2}\tilde{A}\tilde{D}^{-1/2}$ and $\tilde{D}^{-1}\tilde{A}$. 
Interestingly, we can prove that graph matching with the random-weight GNN approximates the weight-free GNN under the BasicTFGM framework (proof in Appendix~\ref{sec:proofs}). The intuition is that the projection of a random matrix $W\sim \frac{1}{\sqrt{d}}\mathcal{N}(\mathbf{0}, \mathbf{I})$ can approximately preserve the distance between input vectors, which is well-known as the JL lemma~\cite{johnson1984extensions, shi2012margin}.

\begin{table*}[htbp]
    \centering
    \footnotesize
    \setlength{\tabcolsep}{1.8pt}
    \caption{Accuracy (\%) of keypoint matching on PascalVOC. * denotes results from the original paper~\cite{fey2020deep}.}
    \resizebox{0.99\textwidth}{!}{
    \begin{tabular}{lcccccccccccccccccccc|c}
    \toprule
    \textbf{Methods} & \textbf{Aero} & \textbf{Bike} & \textbf{Bird} & \textbf{Boat} & \textbf{Bot.} & \textbf{Bus} & \textbf{Car} & \textbf{Cat} & \textbf{Cha.} & \textbf{Cow} & \textbf{Tab.} & \textbf{Dog} & \textbf{Hor.} & \textbf{MBike} & \textbf{Per.} & \textbf{Plant} & \textbf{Sheep} & \textbf{Sofa} & \textbf{Train} & \textbf{TV} & \textbf{Mean} \\
    \midrule
    NodeMatch & 21.9 & 27.0 & 30.6 & 39.2 & 37.5 & 66.7 & 54.3 & 38.9 & 19.3 & 34.4 & 78.4 & 29.0 & 49.4 & 30.4 & 30.9 & 40.8 & 36.6 & 81.5 & 52.1 & 70.1 & 43.4 \\
    MLP* & 34.3 & 45.9 & 37.3 & 47.7 & 53.3 & 75.2 & 64.5 & 61.7 & 27.7 & 40.5 & 85.9 & 46.6 & 50.2 & 39.0 & 37.3 & 58.0 & 49.2 & 82.9 & 65.0 & 74.2 & 53.8 \\
    GraphSAGE & 33.8 & 39.8 & 36.6 & 53.1 & 54.9 & 79.7 & 64.2 & 49.6 & 28.6 & 49.1 & 83.4 & 41.9 & 56.3 & 34.5 & 37.6 & 62.0 & 46.7 & 81.4 & 65.4 & 77.5 & 53.8 \\
    SplineCNN* & 42.1 & 57.5 & 49.6 & 59.4 & 83.8 & 84.0 & 78.4 & 67.5 & 37.3 & 60.4 & 85.0 & 58.0 & 66.0 & 54.1 & 52.6 & 93.9 & 60.2 & 85.6 & 87.8 & 82.5 & 67.3 \\
    DGMC* & 47.0 & 65.7 & \textbf{56.8} & \textbf{67.6} & 86.9 & \textbf{87.7} & \textbf{85.3} & \textbf{72.6} & 42.9 & 69.1 & 84.5 & \textbf{63.8} & \textbf{78.1} & 55.6 & \textbf{58.4} & \textbf{98.0} & \textbf{68.4} & 92.2 & 94.5 & \textbf{85.5} & 73.0 \\
    \midrule
    \multicolumn{2}{l}{\textbf{TFGM}} & & & & & & & & & & & & & & & & & & & & \\
    GraphSAGE & 25.6 & 30.0 & 31.3 & 44.0 & 38.7 & 70.9 & 54.7 & 41.2 & 21.9 & 33.0 & 80.7 & 29.8 & 47.3 & 28.5 & 30.7 & 48.2 & 39.4 & 83.1 & 60.0 & 74.0 & 45.7 \\
    SplineCNN & 25.9 & 37.7 & 38.4 & 58.3 & 68.0 & 83.2 & 70.1 & 48.1 & 29.3 & 43.5 & 82.8 & 37.3 & 59.6 & 37.6 & 38.4 & 73.9 & 44.1 & 93.6 & 79.1 & 80.3 & 56.5 \\
    DGMC & 27.9 & 39.6 & 43.4 & 63.3 & 78.3 & 85.3 & 76.6 & 55.1 & 31.4 & 47.2 & 85.8 & 41.2 & 62.9 & 36.4 & 53.1 & 86.0 & 46.0 & 96.0 & 88.0 & 83.3 & 61.3 \\
    \midrule
    \multicolumn{2}{l}{\textbf{TFGMws}} & & & & & & & & & & & & & & & & & & & & \\
    GraphSAGE & 39.4 & 43.6 & 42.2 & 48.4 & 57.4 & 74.9 & 59.9 & 53.5 & 26.7 & 39.6 & 78.9 & 39.2 & 56.6 & 42.9 & 34.3 & 65.9 & 42.4 & 87.8 & 65.0 & 76.6 & 53.8 \\
    SplineCNN & 48.2 & 65.0 & 49.2 & 61.2 & 84.9 & 83.6 & 80.4 & 62.3 & 50.1 & 62.0 & 86.4 & 54.5 & 67.6 & 64.4 & 53.0 & 97.0 & 58.3 & 97.1 & 93.7 & 84.7 & 70.2 \\
    DGMC & \textbf{53.9} & \textbf{72.1} & 56.6 & \textbf{67.6} & \textbf{87.7} & 86.5 & 84.9 & 68.3 & \textbf{56.1} & \textbf{72.1} & \textbf{90.1} & 58.5 & 74.2 & \textbf{70.5} & 58.2 & 97.4 & 62.2 & \textbf{97.5} & \textbf{95.2} & 85.2 & \textbf{74.7} \\
    \bottomrule
    \end{tabular}%
    }
    \label{tab:pascal}%
    % \vspace{-0.2cm}
\end{table*}%

The analysis above indicates that random-weight GNN should perform slightly worse than the weight-free GNN, which is demonstrated in experiments (Appendix~\ref{appendix:experiment}). Thus, we suggest the weight-free than the random-weight GNN.

\subsection{Generalizing to Different Graph Operators (P4)}
The adjacency matrix $A$ in Equation~(\ref{eq:gnn}) is a graph operator, which defines how GNN aggregate message from neighboring nodes. Let $\tilde{A}=A+I$ and $\tilde{D}_{ii}=\sum_{j}\tilde{A}_{ij}$. Proposition~\ref{proposition:1} and Proposition~\ref{proposition:2} can be generalized to other GCN-flavored GNNs like GCN~\cite{kipf2016semi} and GraphSAGE~\cite{hamilton2017inductive} when using the corresponding graph operator in the graph matching objective:
\begin{equation*}
 S^* = \argmax_{S\in \Tcal} \sum_{\substack{i\in \mathcal{V}^{(s)}\\ j\in \mathcal{V}^{(t)}}}Q_{ij}S_{ij} + \sum_{\substack{i,i'\in \mathcal{V}^{(s)}\\ j,j'\in \mathcal{V}^{(t)}}}\hat{A}_{i,i'}^{(s)}\hat{A}_{j,j'}^{(t)}P_{i'j'}^{(ij)}S_{ij},
 \vspace{-0.2cm}
\end{equation*}
where $\hat{A} = \tilde{D}^{-1/2}\tilde{A}\tilde{D}^{-1/2}$ for GCN; $\hat{A}=\tilde{D}^{-1}\tilde{A}$ for GraphSAGE. Compared to Equation~(\ref{eq:new_gm}), the only difference is that the previous binary indicator of edge existence is now scaled by node degrees. By this generalization, we can apply TFGM with the popular GCN and GraphSAGE.

\textbf{Generalizing to More Powerful GNNs}. This TFGM approach is not limited by the GCN-flavored architecture for its principles are universal priors of graph matching. We have experimentally verified that other GNNs, \textit{e.g.}, SplineCNN ~\cite{fey2018splinecnn}, also perform promisingly under TFGM. For further exploration, we feel more powerful graph operators~\cite{corso2020principal} and representing semantic edge features~\cite{gilmer2017neural} are promising directions for training-free graph matching.

% In real application, we are free to explore more powerful graph operators and kernels without being limited by the QAP formulation.}

%% file: experiment.tex
\section{Experiments}
\label{sec:experiment}
We experiment on three benchmarks under supervised, semi-supervised, and unsupervised settings. The experimental purpose is not to surpass the fully trained state-of-the-art, but to verify TFGM's superiority for boosting GNNs' performances under the training-free setting and provide empirical evidence of our analysis. 
% We show that TFGM with three widely used GNNs can achieve competitive performance to their fully trained counterparts and other strong baselines. 
We also conduct ablation studies to test TFGM's key components and efficiency.

\textbf{Experimental Setup}.
We use GraphSAGE~\cite{hamilton2017inductive}, SplineCNN\cite{fey2018splinecnn}, and DGMC~\cite{fey2020deep} as the baseline GNNs for TFGM due to their popularity in graph matching. To investigate the effectiveness of training and utilizing annotation (\textbf{P2}), we present GNNs' performances of three versions: the original \textbf{fully trained version}, \textbf{TFGM}, and \textbf{TFGMws}. Note that the models in the following tables without TFGM or TFGMws are fully trained. We use the same LAP solver as the baselines in experiments for fair comparison. The detailed experimental setup is in Appendix~\ref{sec:exp_setup}. 

% \textbf{Baseline GNNs}.  We remove the nonlinear activations between layers and use the weight-free version of GraphSAGE. 
% Details are in Appendix~\ref{sec:baseline_gnns}.

\subsection{Supervised Keypoint Matching on Images}
Keypoint matching is supervised graph matching to find the semantic equivalent keypoints between images of the same objects. We experiment on PascalVOC~\cite{everingham2010pascal} with Berkeley annotation~\cite{bourdev2009poselets}, a benchmark with $20$ categories of objects (Table~\ref{tab:pascal}).

First, TFGMws performs better than fully trained GNNs on mean accuracy. This demonstrates training-free methods' ability to integrate node features and structure features. We attribute TFGMws' better performance to the preserved neighborhood at different localities. TFGMws strictly compares nodes' neighbors within the same order, while trained GNNs fail to guarantee such similarity measurement.
Second, TFGMws shows significant improvement ($11.7\%$ on average) over TFGM. This implies the importance of annotation, and our proposed method can effectively utilize annotation without training. 
% Meanwhile, \textit{GraphSAGE with TFGM} achieves limited performance gains over NodeMatch, because in PascalVOC, graphs are small (at most $19$ nodes) and can only provide limited structural information for matching.
Third, TFGM benefits from advanced architectures: DGMC and SplineCNN significantly outperform GraphSAGE. This coincidences with \textbf{P4} that TFGM can capture the inductive bias in GNNs. 
% More importantly, this indicates the potentials to improve TFGM via more sophisticated neural architectures.

\begin{table}[t]
 \footnotesize
 \centering
 \caption{Entity alignment performance on DBP15k. * indicates performance from original papers.}
 {
 \begin{tabular}{lccccccccc}
 \toprule
 & \multicolumn{3}{c}{\textbf{ZH-EN}} & \multicolumn{3}{c}{\textbf{JA-EN}} & \multicolumn{3}{c}{\textbf{FR-EN}} \\
 \textbf{Methods} & \textbf{H@1} & \textbf{H@10} & \textbf{MRR} & \textbf{H@1} & \textbf{H@10} & \textbf{MRR} & \textbf{H@1} & \textbf{H@10} & \textbf{MRR} \\
 \midrule
 NodeMatch & 60.3 & 71.1 & 0.641 & 66.6 & 77.1 & 0.704 & 84.6 & 91.2 & 0.871 \\
 MLP & 61.1 & 69.8 & 0.643 & 68.7 & 77.9 & 0.721 & 89.0 & 93.9 & 0.909 \\
 EVA* & 76.1 & \textbf{90.7} & 0.814 & 76.2 & 91.3 & 0.817 & 79.3 & 94.2 & 0.847 \\
 NMN* & 73.3 & 86.9 & - & 78.5 & 91.2 & - & 90.2 & 96.7 & - \\
 GraphSAGE & 69.4 & 85.1 & 0.752 & 73.8 & 88.3 & 0.790 & 87.9 & 95.8 & 0.909 \\
 DGMC* & 80.1 & 87.5 & - & 84.8 & 89.7 & - & 93.3 & 96.0 & - \\
 \midrule
 \textbf{TFGM} & & & & & & & & & \\
 GraphSAGE & 66.9 & 79.1 & 0.712 & 74.2 & 85.1 & 0.781 & 88.6 & 94.6 & 0.908 \\
 DGMC & 79.1 & 84.5 & 0.811 & 84.6 & 90.5 & 0.869 & 94.4 & 97.0 & 0.954 \\
 \midrule
 \textbf{TFGMws} & & & & & & & & & \\
 GraphSAGE & 69.0 & 81.1 & 0.733 & 75.7 & 86.5 & 0.795 & 89.2 & 95.2 & 0.914 \\
 DGMC & \textbf{81.4} & 86.3 & \textbf{0.833} & \textbf{86.2} & \textbf{91.5} & \textbf{0.883} & \textbf{94.9} & \textbf{97.3} & \textbf{0.959} \\
 \bottomrule
 \end{tabular}%
 }
 \label{tab:dbp15k}%
 \vspace{-0.2cm}
\end{table}% 

\begin{table}[t]
   \footnotesize
   \centering
   \caption{Accuracy (\%) of Network Alignment on the PPI dataset. We report the baseline performances from our re-implementation with their released source code.}
   % \resizebox{0.48\textwidth}{!}{
   % \setlength{\tabcolsep}{1mm}
   {
   \begin{tabular}{lcccccccccc}
   \toprule
   & \multicolumn{5}{c}{\textbf{Low-conf. Edges}} & \multicolumn{5}{c}{\textbf{Random Rewirement}} \\
   \textbf{Noise Ratio} & \textbf{5\%} & \textbf{10\%} & \textbf{15\%} & \textbf{20\%} & \textbf{25\%} & \textbf{5\%} & \textbf{10\%} & \textbf{15\%} & \textbf{20\%} & \textbf{25\%} \\
   \midrule
   GHOST & 73.6 & 48.9 & 36.7 & 25.5 & 20.4 & 41.7 & 14.9 & 11.6 & ~9.2 & ~7.5 \\
   KerGM & 64.2 & 48.6 & 39.7 & 30.7 & 29.9 & 38.6 & 14.6 & ~5.8 & ~1.7 & ~1.0 \\
   MAGNA++ & 74.0 & 66.3 & 58.4 & 49.8 & 43.3 & 67.2 & 44.6 & 36.9 & 34.1 & 30.6 \\
   \midrule
   \multicolumn{3}{l}{\textbf{TFGM one-hot}} & & & & & & & & \\
   GraphSAGE & 65.1 & 46.2 & 32.1 & 29.1 & 23.5 & 83.3 & 81.1 & \textbf{78.1} & \textbf{75.0} & \textbf{71.7} \\
   DGMC & 79.0 & 74.2 & 40.5 & 48.3 & 34.0 & \textbf{83.6} & \textbf{81.3} & 77.9 & 74.0 & 67.8 \\
   \midrule
   \multicolumn{3}{l}{\textbf{TFGM pos-enc}} & & & & & & & & \\
   GraphSAGE & 79.9 & 62.6 & 52.7 & 41.1 & 33.2 & 75.5 & 67.1 & 59.5 & 53.2 & 45.3 \\
   DGMC & \textbf{83.6} & \textbf{78.6} & \textbf{72.4} & \textbf{62.4} & \textbf{50.1} & 81.9 & 76.8 & 68.4 & 58.1 & 50.2 \\
   \bottomrule
   \end{tabular}
   }%
   \label{tab:ppi}%
   \vspace{-0.4cm}
  \end{table}%

\subsection{Semi-supervised Entity Alignment on KGs}
Entity alignment is semi-supervised graph matching aiming to find equivalent entities between two KGs. We use a benchmark DBP15k~\cite{sun2017cross} with KGs between language pairs: Chinese to English (ZH-EN), Japanese to English (JA-EN), and French to English (FR-EN). We compare with unsupervised method EVA~\cite{liu2020visual} and GNN model NMN~\cite{wu2020neighborhood}. Table~\ref{tab:dbp15k} presents performances of Hit@$1$ (H@$1$), Hit@$10$ (H@$10$) and Mean Reciprocal Rank (MRR). 

First, \textit{DGMC with TFGMws} outperforms its fully trained counterpart on $5$ out of $6$ metrics, demonstrating training-free methods' ability to integrate node and structure features.
% This is because DBP15k has large graphs with more than $10k$ nodes and $50k$ edges. These luxuriant structures provide discriminative matching signals as a complement to node features.
Second, TFGM significantly outperforms the unsupervised baseline EVA and fully trained baseline NMN while enjoying an efficient training-free property. This performance demonstrates TFGM's superiority in the unsupervised setting. Note that EVA uses extra visual features, which significantly boost the Hit@$10$ scores. 
% This suggests that stronger node features can further improve TFGM's performance.
Third, TFGMws outperforms TFGM by only $1.4\%$ Hit@$1$ on average. Compared with supervised graph matching (Table~\ref{tab:pascal}), the improvement is limited. We attribute this to the weak supervision in DBP15k: the unsupervised baseline EVA's performance is very competitive to supervised baselines. In addition, shown by the decent performance of NodeMatch, the initial node features are strong for measuring semantic equivalence. 
%This shows the significance of TFGM for transferring advanced GNN architectures to areas that training-based models cannot previously attain.
% We can see GraphSAGE outperforms MLP by a large margin, unlike their similar performance on PascalVOC.

\subsection{Unsupervised Network Alignment of Protein-Protein Interaction Networks}
Protein-Protein Interaction (PPI) Network Alignment is unsupervised and aims to find corresponding proteins in networks of different species. We use the Low-conf. Edges dataset~\cite{collins2007toward} and Random Rewirement dataset~\cite{saraph2014magna} for evaluation. Each dataset has $5$ versions of different noise ratios. We compare with unsupervised baselines GHOST~\cite{patro2012global}, KerGM~\cite{zhang2019kergm}, and MAGNA++~\cite{vijayan2015magna++}. This dataset has no node features. We thus initialize the nodes with the one-hot encoding or positional-encoding~\cite{vaswani2017attention} of node degrees. Table~\ref{tab:ppi} reports the matching accuracy. 

First, we can see that TFGM outperforms baselines by a large margin ($17.7\%$ on average), demonstrating the superiority of TFGM in the unsupervised setting. We attribute the good performance to TFGM's effectiveness in capturing high-
\begin{figure}[t]
   \begin{subfigure}[b]{0.475\textwidth}
      \captionsetup{size=footnotesize, labelfont=footnotesize}
   \includegraphics[width=\textwidth]{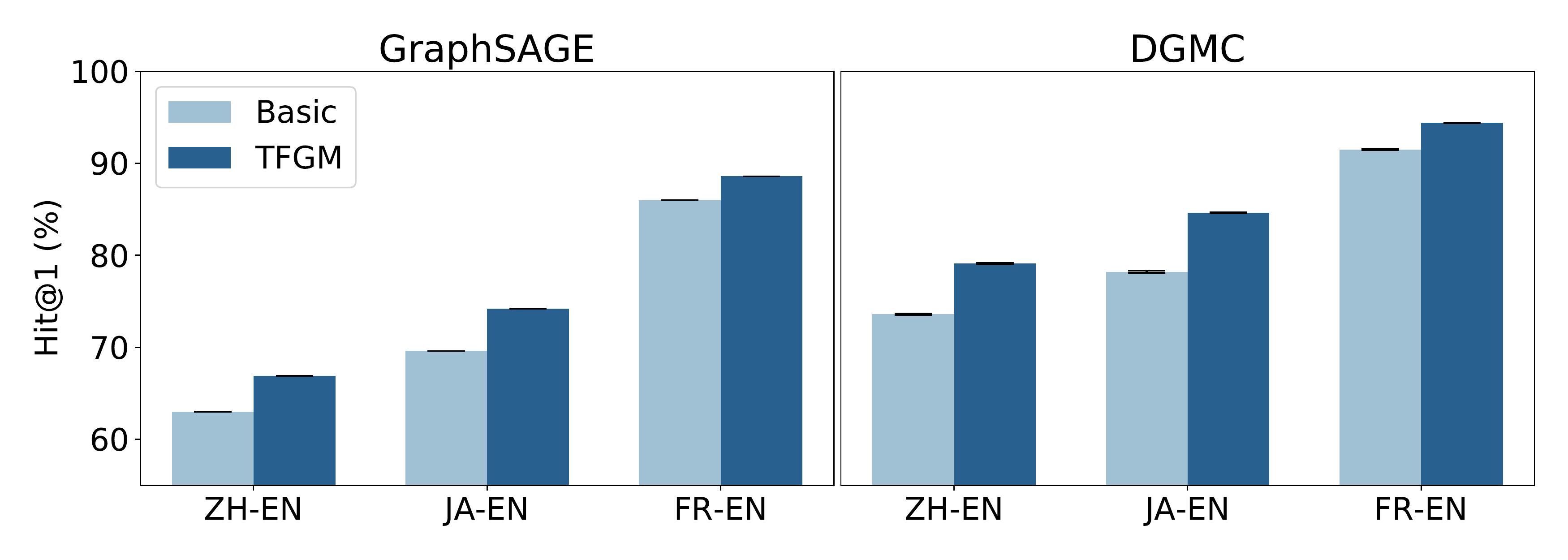}
   \caption{Training-free GNNs' performance on DBP15k.}
   \end{subfigure}
   \begin{subfigure}[b]{0.475\textwidth}
      \captionsetup{size=footnotesize, labelfont=footnotesize}
   \includegraphics[width=\textwidth]{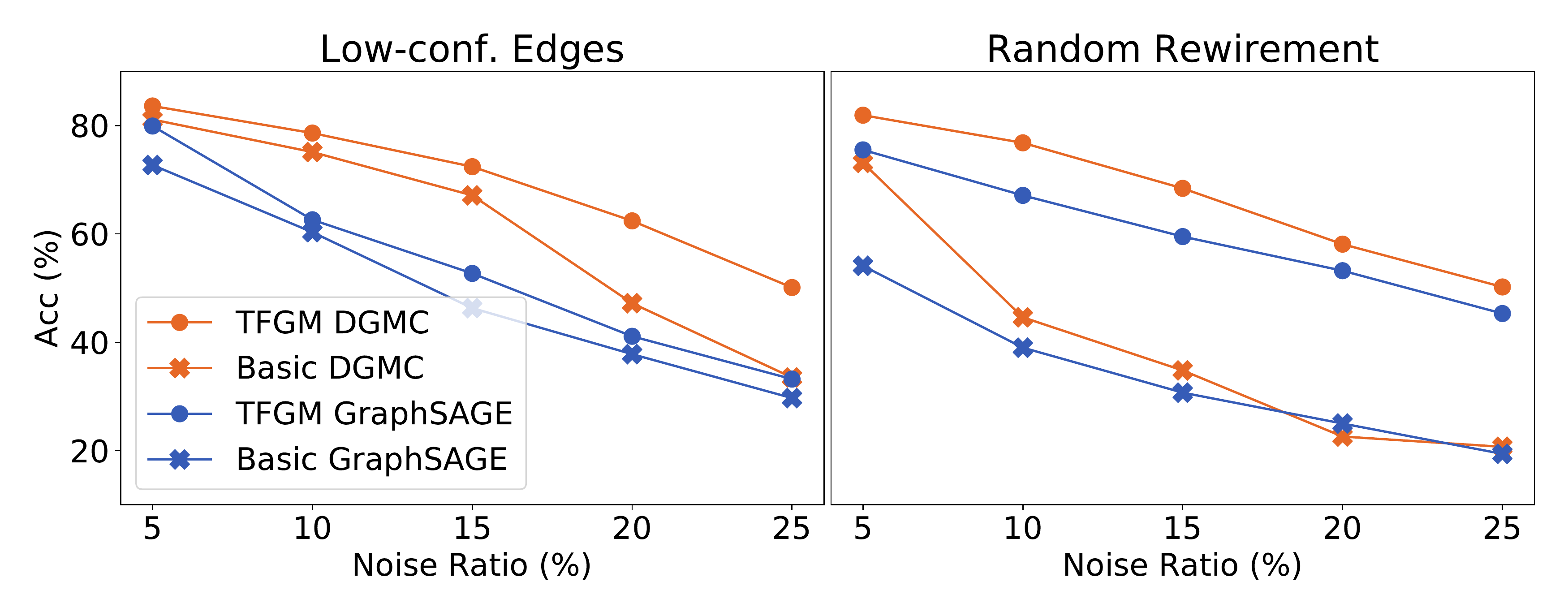}
   \caption{Training-free GNNs with positional encoding on PPI.}
   \end{subfigure}
   \caption{Comparison of the TFGM and the BasicTFGM.}
   \label{fig:tfgm_vs_basic}
   \vspace{-0.3cm}
\end{figure}

% \begin{wrapfigure}[]{l}[0pt]{0.5\textwidth}
%    \centering
%    \scriptsize
%    \makeatletter\def\@captype{table}\makeatother
%    \caption{Average running time (seconds) of 5 independent runs for the best performing baselines, TFGM, and TFGMws. We do not count the time for loading datasets.}
%    \begin{tabular}{lccc}
%    \toprule
%    \multicolumn{1}{l}{\textbf{Dataset}} & \textbf{BestBaseline} & \textbf{TFGM} & \textbf{TFGMws} \\
%    \midrule
%    PascalVOC            & $504.3$        & $18.1$ & $67.2$ \\
%    DBP15k               & $44.1$         & $5.5$  & $5.7$  \\
%    PPI                  & $983.0$        & $0.2$  & -   \\ 
% \bottomrule
% \end{tabular}%
% \label{tab:abl_time}%
% \end{wrapfigure}
   
\begin{wrapfigure}[35]{r}[0pt]{0.5\textwidth}
   \centering
   \footnotesize
   \setlength{\tabcolsep}{1.5mm}
   \makeatletter\def\@captype{table}\makeatother
   \caption{Average running time (seconds) of 5 independent runs for the best performing baselines, TFGM, and TFGMws. We do not count the time for loading datasets.}
   \begin{tabular}{lccc}
   \toprule
   \multicolumn{1}{l}{\textbf{Dataset}} & \textbf{BestBaseline} & \textbf{TFGM} & \textbf{TFGMws} \\
   \midrule
   PascalVOC            & $504.3$        & $18.1$ & $67.2$ \\
   DBP15k               & $44.1$         & $5.5$  & $5.7$  \\
   PPI                  & $983.0$        & $0.2$  & -   \\ 
   \bottomrule
   \end{tabular}%
   \label{tab:abl_time}%

   % % % % % % % % % % % % % % % % % % % % % 
    \begin{subfigure}[b]{0.475\textwidth}
      \captionsetup{size=footnotesize, labelfont=footnotesize}
    \includegraphics[width=\textwidth]{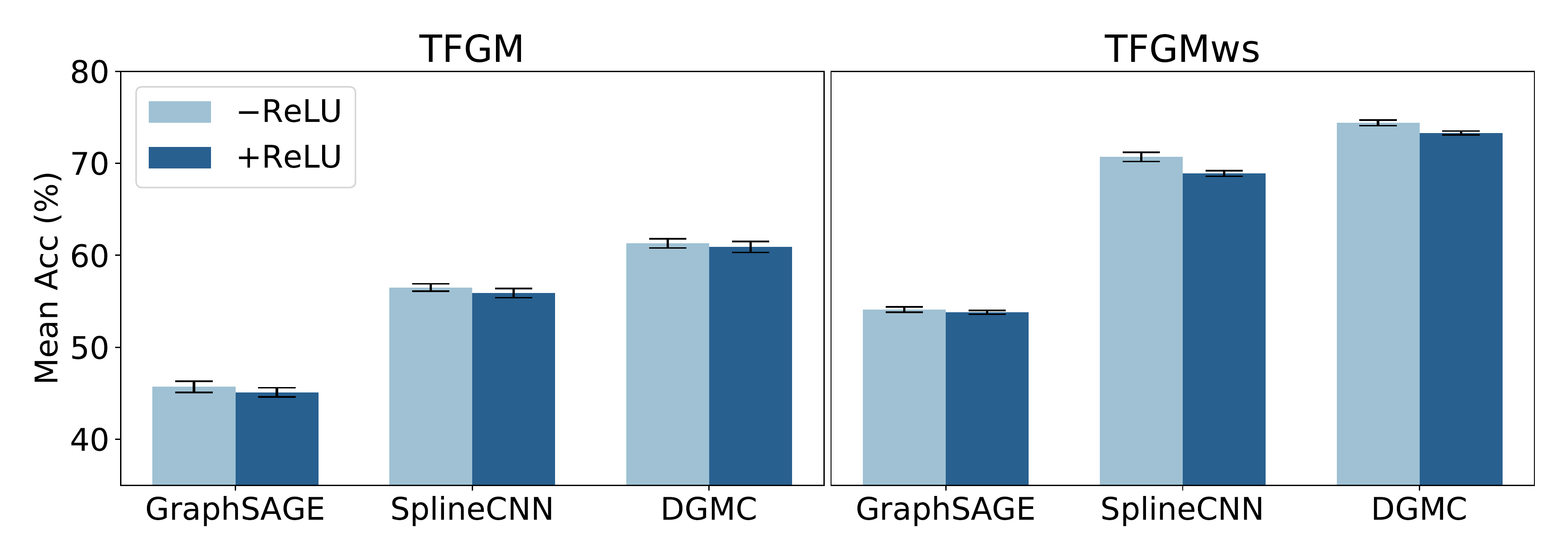}
    \caption{TFGM with GNNs on PascalVOC.}
    \end{subfigure}
    \begin{subfigure}[b]{0.475\textwidth}
      \captionsetup{size=footnotesize, labelfont=footnotesize}
    \includegraphics[width=\textwidth]{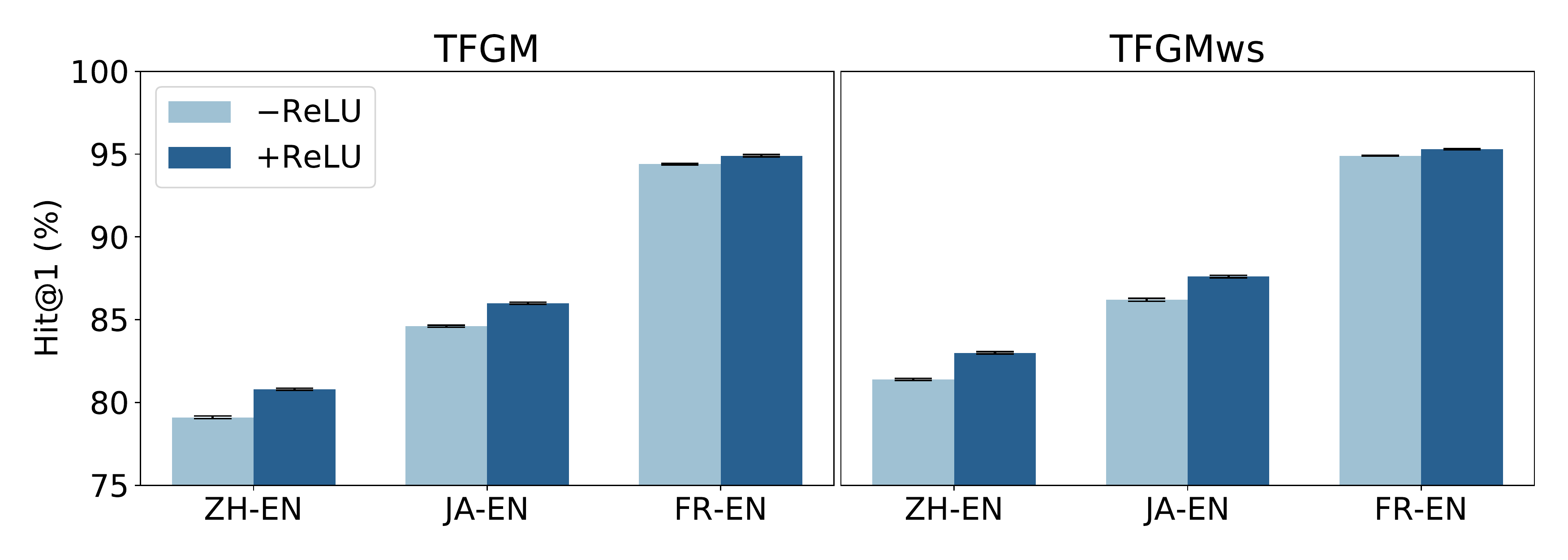}
    \caption{TFGM with DGMC on DBP15k.} %Omit GraphSAGE to save space.} 
    \end{subfigure}
    \begin{subfigure}[b]{0.475\textwidth}
      \captionsetup{size=footnotesize, labelfont=footnotesize}
    \includegraphics[width=\textwidth]{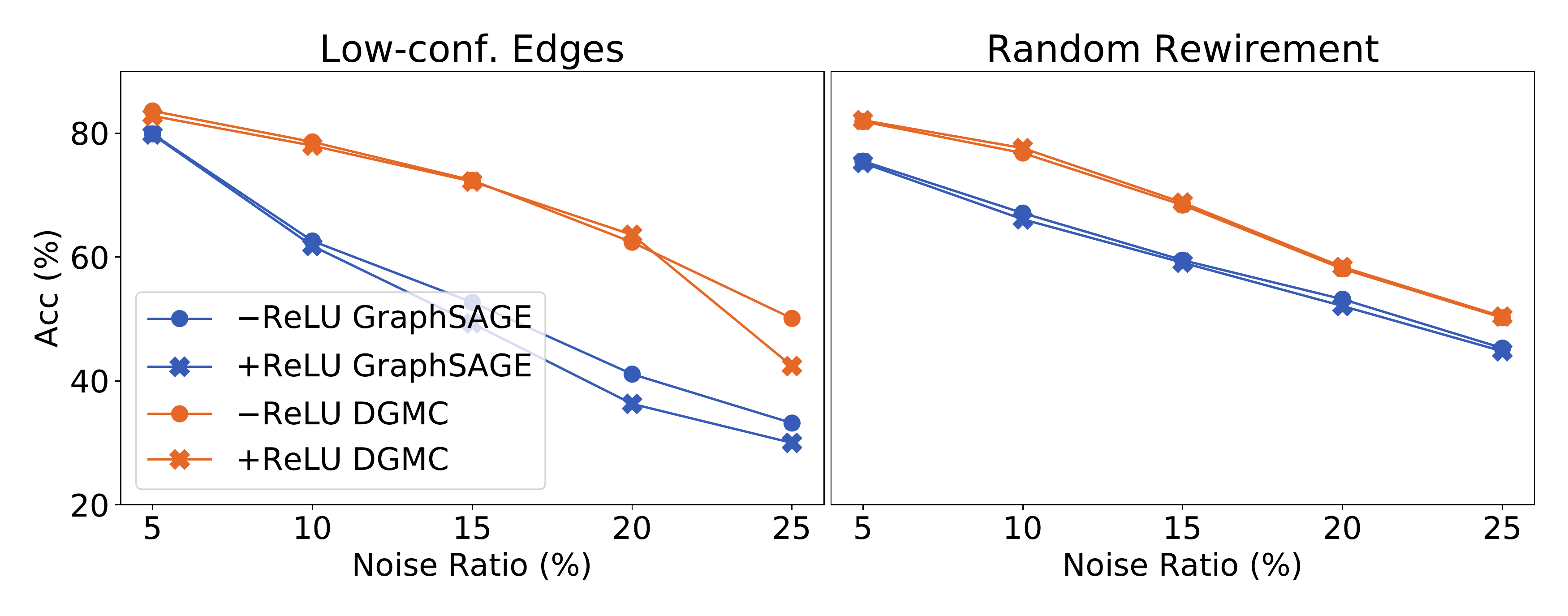}
    \caption{TFGM pos-enc with GNNs on PPI.}
    \end{subfigure}
    \makeatletter\def\@captype{figure}\makeatother
    \caption{TFGM's performance with and without ReLU.}
    \label{fig:linear_nonlinear}
\end{wrapfigure}%

order neighborhoods with deep GNNs (\textit{i.e.}, $10$ layers). Baselines use only one-hop neighbors of nodes. % and one-hop edges. 
Second, TFGM one-hot performs best among models on the Random Rewirement dataset. This is because this dataset is generated while ensuring equivalent nodes to have the same degree, which provides a strong bias for one-hot matching. 
% On the other hand, in the more realistic Low-conf. Edges dataset without degree guarantee, DGMC with positional encoding performs better. 
As a note, NodeMatch that compares only degree features has accuracy $<10\%$ due to duplicate degrees.
Third, GraphSAGE outperforms DGMC on the Random Rewirement dataset if using one-hot encoding. This is expected because the ``rewirement'' noise violates the structural consistency, thus confuses DGMC's refinement towards the consistency.
% On the contrary, the noisy edges in the Low-conf. Edges dataset have biological meanings, thus will not hurt DGMC's refinement steps.

\noindent \textbf{Discussion}. PPI dataset has no node features, thus matching relies on the assumption of structural consistency~\cite{wang2018cross}. On the Low-conf. Edges dataset, this assumption holds exactly because the source graph is a subgraph of the target graph.
% on the other dataset, the structural consistency is violated. 
Therefore, optimizing QAP (Equation (\ref{eq:gm})) guarantees the optimal performance on the Low-conf. Edges dataset. Our experiments on it give empirical evaluation for TFGM's approximation of the QAP objective.

\subsection{Ablation Studies}
\label{sec:ablation}
\noindent\textbf{Preserving Neighbors of Different Localities}.
To further demonstrate the importance of preserving neighbors of different orders, we compare training-free GNNs under different frameworks: BasicTFGM based on the $L$-th order neighborhoods and TFGM based on all of $(0, 1,\cdots,L)$-th orders' neighborhoods. For BasicTFGM, we use residual connection~\cite{he2016deep} to preserve the node compatibility. Figure~\ref{fig:tfgm_vs_basic} shows that GNNs under the TFGM framework consistently outperform their BasicTFGM counterparts ($4.3\%$ on average on DBP15k, $16.7\%$ on average on PPI). We observe similar results on PascalVOC (Appendix~\ref{appendix:additional}).

\noindent\textbf{Efficiency}. 
% TFGM is supposed to be more efficient than fully trained models. 
We present the average running time in Table~\ref{tab:abl_time} to evaluate TFGM's efficiency. We compare models with the best performance to obtain the most representative results. For baselines, we use DGMC on PascalVOC and DBP15k and use MAGNA++ on PPI. For our framework, we stick with DGMC as the backbone GNN. 

First, we highlight that our framework performs empirically faster than all baselines by a large margin.
On PascalVOC and DBP15k, TFGMws is $7\sim 8$ times faster than the fully trained models due to the cut-down of the training phase. On PPI, TFGM is nearly $5000$ times faster than MAGNA++ due to TFGM's efficiency and GPU acceleration. Second, TFGM further improves over TFGMws on speed for not relying on training data. Considering TFGM's better performance than some supervised models in experiments (Table~\ref{tab:pascal},~\ref{tab:dbp15k}), it makes a fast promising solution for graph matching. 
Third, the gap between TFGM and TFGMws is small on DBP15k, showing that our strategy for utilizing annotation introduces little computation overhead in the semi-supervised setting. Complexity analysis is in Appendix~\ref{appendix:complexity}.

\noindent\textbf{Removing Nonlinear Activation}. 
We now support our conjecture that \textit{it is inessential to use the nonlinear activation in training-free GNNs for graph matching} (Section~\ref{sec:rand_vs_free}) with empirical evidence. In Figure~\ref{fig:linear_nonlinear}, we compare TFGM's performance with and without the ReLU nonlinearity, the most common activation function. ReLU's contribution to performance is unstable across datasets: on PascalVOC, without ReLU is better; on DBP15k, with ReLU is better; on PPI, ReLU makes no significant difference. We conclude that ReLU is inessential in the training-free setting.

%% file: appendix.tex
\begin{center}
  {\titlefont \textbf{Supplemetary Material for Training-Free Graph Neural Networks for Graph Matching}}
\end{center}
\vspace{3em}
\section{Proofs}
\label{sec:proofs}
Proof of Proposition~\ref{proposition:1}:
\begin{proof}
Let $\phi(\mathcal{G})=\mathrm{GNN}_L(A, X)$, we need to show that Eq.~\ref{eq:objective} can be rewritten as Eq.~\ref{eq:new_gm}:
{\footnotesize
\begin{align*}
  S^* & = \argmax_{S\in \Tcal} \sum_{\substack{i\in \mathcal{V}^{(s)}\\ j\in \mathcal{V}^{(t)}}} S_{ij}\left(\mathrm{GNN}_L(A^{(s)}, X^{(s)})\mathrm{GNN}_L(A^{(t)}, X^{(t)})\T\right)_{ij} \\
  & = \argmax_{S\in \Tcal} \sum_{\substack{i\in \mathcal{V}^{(s)}\\ j\in \mathcal{V}^{(t)}}} S_{ij}\left((A^{(s)} (H^{(L-1)})^{(s)}W_L)(A^{(t)} (H^{(L-1)})^{(t)}W_L)\T\right)_{ij}
\end{align*}}

\noindent Substitute $P=(H^{(L-1)})^{(s)}W_L((H^{(L-1)})^{(t)}W_L)\T$:
{\footnotesize
\begin{align*}
S^* & = \argmax_{S\in \Tcal} \sum_{i\in \mathcal{V}^{(s)}, j\in \mathcal{V}^{(t)}} S_{ij}(A^{(s)}P(A^{(t)})\T)_{ij}\\
  & = \argmax_{S\in \Tcal}\sum_{i,i'\in \mathcal{V}^{(s)}, j,j'\in \mathcal{V}^{(t)}}A_{ii'}^{(s)}A_{jj'}^{(t)}S_{ij}P_{i'j'}
\end{align*}}
\end{proof}

Proof of Proposition~\ref{proposition:2}:
\begin{proof}
Let $g_l(\mathcal{G})=\mathrm{GNN}_L(A,X)$, we need to rewrite Equation~(\ref{eq:cos_gm}) as Equation~(\ref{eq:new_gm}). 

For any matrix $B \in \RR^{n\times m}$, we define $B_{i:} \in \RR^{1\times m}$ to be its $i$-th row vector for $i=1,\dots,n$. Let $M^{(l)}_{ij}=1/Z^{(l)}_{ij}$.  Let $S^*$ be the solution of \textit{TFGM with $\mathrm{GNN}_L(A, X)$}. Then, by the definition of the cosine similarity and the linearity of matrix indexing, 
{\footnotesize
\begin{align*}
S^* &= \argmax_{S\in \Tcal} \sum_{\substack{i\in \mathcal{V}^{(s)}\\ j\in \mathcal{V}^{(t)}}}S_{ij}\left(\sum_{l=0}^L \mathrm{Cos}\left(\mathrm{GNN}_{l}(A^{(s)}, X^{(s)}), \mathrm{GNN}_{l}(A^{(t)}, X^{(t)})\right)\right)_{ij}
\\ & =\argmax_{S\in \Tcal} \sum_{\substack{i\in \mathcal{V}^{(s)}\\ j\in \mathcal{V}^{(t)}}}S_{ij}\sum_{l=0}^L\frac{\langle\mathrm{GNN}_{l}(A^{(s)}, X^{(s)})_{i:}, \mathrm{GNN}_{l}(A^{(t)}, X^{(t)})_{j:}\rangle}{M^{(l)}_{ij}}
\end{align*}} 
For the $l=0$ term,
{\footnotesize
\begin{align*}
\frac{\langle\mathrm{GNN}_{0}(A^{(s)}, X^{(s)})_{i:}, \mathrm{GNN}_{0}(A^{(t)}, X^{(t)})_{j:}\rangle}{M^{(0)}_{ij}}=\frac{\langle X^{(s)}_{i:}, X^{(t)}_{j:} \rangle}{M^{(0)}_{ij}} = Q_{ij}.
\end{align*}}
For the $l \ge 1$ term,
{\footnotesize
\begin{align*}
&\sum_{l=1}^{L}\frac{\langle A^{(s)}_{i:}(H^{(l-1)})^{(s)}W_l, A^{(t)}_{j:} (H^{(l-1)})^{(t)}W_l\rangle}{M^{(l)}_{ij}}
\\ &=\sum_{l=1}^L\frac{\sum_{i'\in \mathcal{V}^{(s)}, j'\in \mathcal{V}^{(t)}} A^{(s)}_{ii'} A^{(t)}_{jj'} \left((H^{(l-1)})^{(s)}W_l((H^{(l-1)})^{(t)}W_l)\T \right)_{i'j'} }{M^{(l)}_{ij}}
\\ & =\sum_{\substack{i'\in \mathcal{V}^{(s)}\\ j'\in \mathcal{V}^{(t)}}}A^{(s)}_{ii'} A^{(t)}_{jj'}\left(\sum_{l=1}^L \frac{1}{M^{(l)}_{ij}} (H^{(l-1)})^{(s)}W_l((H^{(l-1)})^{(t)}W_l)\T \right)_{i'j'}
\\ & =\sum_{\substack{i'\in \mathcal{V}^{(s)}\\ j'\in \mathcal{V}^{(t)}}}A^{(s)}_{ii'} A^{(t)}_{jj'}P_{i'j'}^{(ij)} 
\end{align*}}
By combining those, we have that
{\footnotesize
$$
S^*=\argmax_{S\in \Tcal} \sum_{i\in \mathcal{V}^{(s)}, j\in \mathcal{V}^{(t)}}S_{ij} Q_{ij} + \sum_{\substack{i,i'\in \mathcal{V}^{(s)}\\ j,j'\in \mathcal{V}^{(t)}}}A^{(s)}_{ii'} A^{(t)}_{jj'}P_{i'j'}^{(i,j)}S_{ij},
$$}
which proves the desired statement.
\end{proof}

Proof of that graph matching with the random-weight GNN approximates the weight-free GNN under the BasicTFGM framework.
% Proof of that $P_1=\mathbb{E}_{\{W_l\}_{l=1}^L\sim \frac{1}{\sqrt{d}}\mathcal{N}(\mathbf{0}, \mathbf{I})}[P_0]$:
\begin{proof}
Based on Proposition~\ref{proposition:1}, it is easy to show that the objective of \textit{BasicTFGM with random-weight GNN and weight-free GNN} can also be aligned with Equation~(\ref{eq:new_gm}). In specific, we have $Q_0=0$ and $P_0=(A^{(s)})^{L-1}X^{(s)}W_1... W_L ((A^{(t)})^{L-1}X^{(t)}W_1...W_L)\T$ for random-weight GNN; and $Q_1=0$ and $P_1=(A^{(s)})^{L-1}X^{(s)}((A^{(t)})^{L-1}X^{(t)})\T$ for weight-free GNN. 

To show that random-weight GNN approximates the weight-free GNN under the BasicTFGM framework, we need to show that $P_0$ is an unbiased estimator of $P_1$: 
\begin{equation*}
P_1=\mathbb{E}_{\{W_l\}_1^L\sim \frac{1}{\sqrt{d}}\mathcal{N}(\mathbf{0}, \mathbf{I})}[P_0],
\end{equation*}

In Section~\ref{sec:problem}, we define weights $\{W_l\}_{1}^L$ to be sampled from $\frac{1}{\sqrt{d}}\mathcal{N}(\mathbf{0}, \mathbf{I})$. Thus, we have:
{\footnotesize
\begin{align}
  \mathbb{E}_{\{W_l\}_{l=1}^L}[P_0] & = \mathbb{E}[((A^{(s)})^{L-1}X^{(s)} W_1 \cdots W_L) ((A^{(t)})^{L-1}X^{(t)}W_1 \cdots W_L)\T] \\
  \label{eq:f_g_hat}
  & = (A^{(s)})^{L-1}X^{(s)} \mathbb{E}[W_1 \cdots W_L W_L\T \cdots W_1\T] ((A^{(t)})^{L-1}X^{(t)})\T
\end{align}}

Let $K_L=W_1 \cdots W_L W_L\T \cdots W_1\T$, we prove by induction that 
\begin{equation}
\label{eq:base}
\mathbb{E}[K_L]=I \text{, for any integer } L>0.
\end{equation}

\underline{Base case}: $L=1$, it is easy to show that $\mathbb{E}[K_1] = \mathbb{E}[W_1 W_1\T] = I$ (using the independence of random variables). 

\underline{Inductive step}: suppose Eq.~\ref{eq:base} holds for $L-1$, \textit{i.e.}, $\mathbb{E}[K_{L-1}]=I$.

Let $B_{L-1} = W_1 ... W_{L-1}$, then we have $K_L = B_{L-1} W_L W_L\T B_{L-1}\T$ and
{\footnotesize
\begin{align*}
  \mathbb{E}[(K_L)_{ij}] & = \mathbb{E}[\sum_k (B_{L-1} W_L)_{ik} (B_{L-1}W_L)_{jk}] \\
  & = \mathbb{E}[\sum_{k,a,b} (B_{L-1})_{ia} (W_L)_{ak} (B_{L-1})_{jb} (W_L)_{bk}] \\
  & = \sum_{k,a,b} \mathbb{E}[(B_{L-1})_{ia} (W_L)_{ak} (B_{L-1})_{jb} (W_L)_{bk}] \\
  & = \sum_{k,a,b} \mathbb{E}[(B_{L-1})_{ia}(B_{L-1})_{jb}] \mathbb{E}[(W_L)_{ak}(W_L)_{bk}]
\end{align*}}
If $i\neq j, a\neq b$,
{\footnotesize
\begin{align*}
  & \mathbb{E}[(B_{L-1})_{ia}(B_{L-1})_{jb}] \mathbb{E}[(W_L)_{ak}(W_L)_{bk}] \\
  = & \mathbb{E}[(B_{L-1})_{ia}(B_{L-1})_{jb}] \mathbb{E}[(W_L)_{ak}]\mathbb{E}[(W_L)_{bk}] = 0
\end{align*}}
If $i\neq j, a=b$,
{\footnotesize
\begin{align*}
  & \mathbb{E}[(B_{L-1})_{ia}(B_{L-1})_{jb}] \mathbb{E}[(W_L)_{ak}(W_L)_{bk}] \\
  = & \mathbb{E}[(B_{L-1})_{ia}(B_{L-1})_{jb}] \mathbb{E}[(W_L)_{ak}^2] = \frac{1}{d}\mathbb{E}[(B_{L-1})_{ia}(B_{L-1})_{jb}]
\end{align*}}
If $i=j, a\neq b$,
{\footnotesize
\begin{align*}
  \mathbb{E}[(B_{L-1})_{ia}(B_{L-1})_{jb}] \mathbb{E}[(W_L)_{ak}(W_L)_{bk}] = 0
\end{align*}}
If $i=j, a=b$,
{\footnotesize
\begin{align*}
  & \mathbb{E}[(B_{L-1})_{ia}(B_{L-1})_{jb}] \mathbb{E}[(W_L)_{ak}(W_L)_{bk}] \\
  = & \mathbb{E}[(B_{L-1})_{ia}^2] \mathbb{E}[(W_L)_{ak}^2] = \frac{1}{d}\mathbb{E}[(B_{L-1})_{ia}^2]
\end{align*}}
Thus, for $K_L$, if $i\neq j$,
{\footnotesize
\begin{align*}
  & \mathbb{E}[(K_L)_{ij}] = \sum_{k,a} \frac{1}{d}\mathbb{E}[(B_{L-1})_{ia}(B_{L-1})_{ja}] \\
  = & \sum_{a}\mathbb{E}[(B_{L-1})_{ia}(B_{L-1})_{ja}] = \mathbb{E}[(K_{L-1})_{ij}] = 0
\end{align*}}
If $i=j$,
{\footnotesize
\begin{align*}
  \mathbb{E}[(K_L)_{ii}] = \sum_{k,a} \frac{1}{d}\mathbb{E}[(B_{L-1})_{ia}^2] = \sum_{a} \mathbb{E}[(B_{L-1})_{ia}^2] = \mathbb{E}[(K_{L-1})_{ii}] = 1
\end{align*}}

We conclude that $\mathbb{E}[K_L] = I$ for all integer $L>0$. Substitute this into Eq.~\ref{eq:f_g_hat}, we have $P_1 = \mathbb{E}_{\{W_l\}_{l=1}^L\sim \frac{1}{\sqrt{d}}\mathcal{N}(\mathbf{0}, \mathbf{I})}[P_0]$.
\end{proof}

\section{Complexity Analysis}
\label{appendix:complexity}
We now analyze the complexity of graph matching using a GCN-flavored weight-free GNN under our framework. 
Let $L$ be the number of GNN layers. $d$ is the dimension of node feature. Given two graphs $\mathcal{G}^{(s)}$ and $\mathcal{G}^{(t)}$, let $\|A\|_0=\mathrm{max}(\|A^{(s)}\|_0, \|A^{(t)}\|_0)$, $|\mathcal{V}|=\mathrm{max}(|\mathcal{V}^{(s)}|, |\mathcal{V}^{(t)}|)$. The complexity analysis is shown in Table~\ref{tab:complexity}. 

\begin{table}[h]
  \centering
  \scriptsize
  \caption{Complexity for graph matching with GCN-flavored weight-free GNN. The underlined part is the complexity for running a Hungarian solver of the LAP, which is optional.}
  \begin{tabular}{lccc}
  \toprule
  \multicolumn{1}{c}{} & Unsupervised & Semi-supervised & Supervised \\
  \midrule
  % Fully trained        & -            &                 &            \\
  TFGM                 & $O(dL|\mathcal{V}|^2) + \underline{O(|\mathcal{V}|^3)}$ & -               & -          \\
  TFGMws               & -            & $O(dL|\mathcal{V}|^2)+ \underline{O(|\mathcal{V}|^3)}$    & $O(dL|\mathcal{V}|\sum_{i=1}^N |\mathcal{V}^{(i)}|) + O(N\log(N)|\mathcal{V}|) + \underline{O(|\mathcal{V}|^3)}$          \\
  \bottomrule
  \end{tabular}
  \label{tab:complexity}
  \end{table}

TFGM's complexity is independent of training dataset size, making it a fast solution for graph matching.

The unsupervised and semi-supervised settings have the same complexity, because our strategy of using semi-supervised annotation is $O(d|\mathcal{I}|)$, where $\mathcal{I}$ is the set of known equivalent node pairs. $O(d|\mathcal{I}|)$ is smaller than $O(d|\mathcal{V}|)$ and thus ignored.

We include the complexity of an optional Hungarian solver of LAP. If Hungarian is used, it dominates the complexity in the unsupervised and semi-supervised setting. However, running GNNs usually costs longer time than running the Hungarian in practice. We can use the less expensive $\argmax$ as LAP solver to achieve $O(dL|\mathcal{V}|^2)$ complexity.

In the supervised setting, we conduct kNN search in a training dataset $\mathcal{D}=\{\mathcal{G}^{(1)}, ..., \mathcal{G}^{(N)}\}$. We assume that the graph matching with the training data uses the $\argmax$ LAP solver for efficiency. The complexity is dominated by two components: 1) graph matching between $\mathcal{G}^{(s)}$ and $\mathcal{G}^{(t)}$ and every graph in the training dataset; 2) sort the similarity scores to find the top k similar nodes. Empirically, the complexity is dominated by the first term $O(dL|\mathcal{V}|\sum_{i=1}^N |\mathcal{V}^{(i)}|)$, which is of the same magnitude or smaller than the fully trained models. However, once being trained, fully trained model performs much faster in the inference phase. We make it a future work to find a more efficient training-free method for the supervised setting.

\section{Experiments}
\label{appendix:experiment}
\subsection{Additional Results}
\label{appendix:additional}
\noindent\textbf{BasicTFGM on PascalVOC}. Table~\ref{tab:pascal_basictfgm} compares the performance of BasicTFGM and TFGM on the supervised graph matching benchmark PascalVOC. TFGM with the three GNNs improves $6.4\%$ on average compared to BasicTFGM. Compare to the BasicTFGM, the only difference is that TFGM concatenates the normalized output from all layers as node embedding. This result again demonstrates the importance of preserving neighbors of all orders for graph matching.

\begin{table*}[h]
\centering
\footnotesize
\setlength{\tabcolsep}{1.25pt}
\caption{Accuracy (\%) of keypoint matching on PascalVOC.}
\resizebox{0.95\textwidth}{!}{
\begin{tabular}{lcccccccccccccccccccc|c}
\toprule
\textbf{Methods} & \textbf{Aero} & \textbf{Bike} & \textbf{Bird} & \textbf{Boat} & \textbf{Bot.} & \textbf{Bus} & \textbf{Car} & \textbf{Cat} & \textbf{Cha.} & \textbf{Cow} & \textbf{Tab.} & \textbf{Dog} & \textbf{Hor.} & \textbf{MBike} & \textbf{Per.} & \textbf{Plant} & \textbf{Sheep} & \textbf{Sofa} & \textbf{Train} & \textbf{TV} & \textbf{Mean} \\
\midrule
\multicolumn{2}{l}{\textbf{BasicTFGM}} & & & & & & & & & & & & & & & & & & & & \\
GraphSAGE & 21.9 & 24.6 & 25.4 & 35.7 & 29.4 & 55.0 & 47.2 & 36.5 & 18.9 & 26.9 & 82.4 & 25.3 & 42.6 & 23.8 & 25.3 & 34.9 & 35.4 & 79.0 & 44.5 & 57.1 & 38.6 \\
SplineCNN & 27.2 & 31.8 & 34.3 & 46.2 & 45.5 & 71.4 & 57.4 & 43.9 & 25.6 & 34.1 & 79.7 & 30.4 & 52.2 & 28.9 & 36.0 & 49.4 & 40.3 & 86.7 & 58.5 & 73.5 & 47.7 \\
DGMC & 30.3 & 42.1 & 43.0 & 56.1 & 65.0 & 82.7 & 71.9 & 53.7 & 27.8 & 43.8 & 84.3 & 38.4 & 64.5 & 40.3 & 46.3 & 70.7 & 48.9 & 93.0 & 76.9 & 81.5 & 58.1 \\
\midrule
\multicolumn{2}{l}{\textbf{TFGM}} & & & & & & & & & & & & & & & & & & & & \\
GraphSAGE & 25.6 & 30.0 & 31.3 & 44.0 & 38.7 & 70.9 & 54.7 & 41.2 & 21.9 & 33.0 & 80.7 & 29.8 & 47.3 & 28.5 & 30.7 & 48.2 & 39.4 & 83.1 & 60.0 & 74.0 & 45.7 \\
SplineCNN & 25.9 & 37.7 & 38.4 & 58.3 & 68.0 & 83.2 & 70.1 & 48.1 & 29.3 & 43.5 & 82.8 & 37.3 & 59.6 & 37.6 & 38.4 & 73.9 & 44.1 & 93.6 & 79.1 & 80.3 & 56.5 \\
DGMC & 27.9 & 39.6 & 43.4 & 63.3 & 78.3 & 85.3 & 76.6 & 55.1 & 31.4 & 47.2 & 85.8 & 41.2 & 62.9 & 36.4 & 53.1 & 86.0 & 46.0 & 96.0 & 88.0 & 83.3 & 61.3 \\  
\bottomrule
\end{tabular}%
}
\label{tab:pascal_basictfgm}%
\end{table*}%

\begin{figure}[h]
\centering
\begin{subfigure}[b]{0.230\textwidth}
    \centering
    \captionsetup{size=footnotesize, labelfont=footnotesize}
    \includegraphics[width=\textwidth]{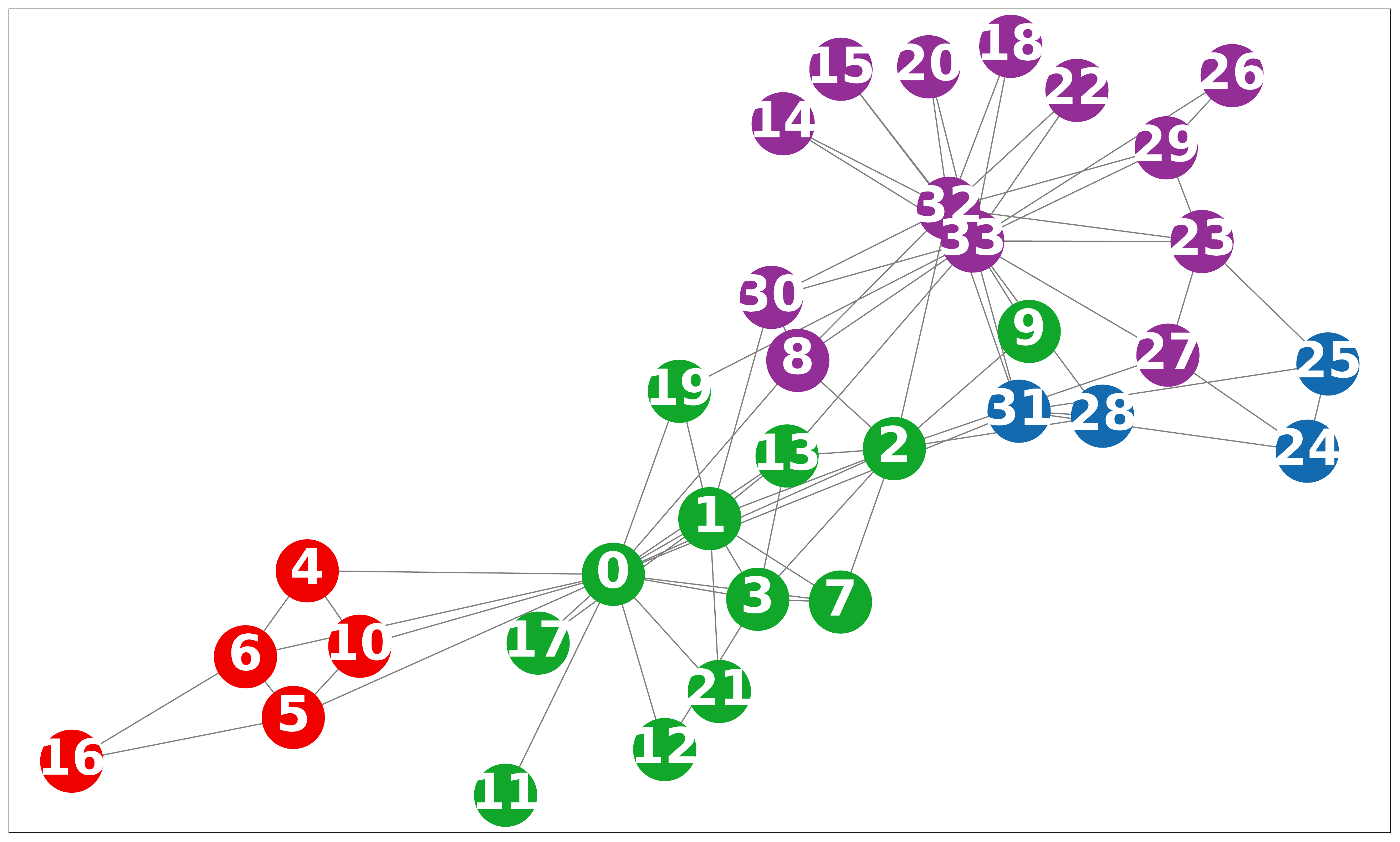}
    \caption{Karate club network.}
\end{subfigure}
\begin{subfigure}[b]{0.230\textwidth}
  \centering
  \captionsetup{size=footnotesize, labelfont=footnotesize}
  \includegraphics[width=\textwidth]{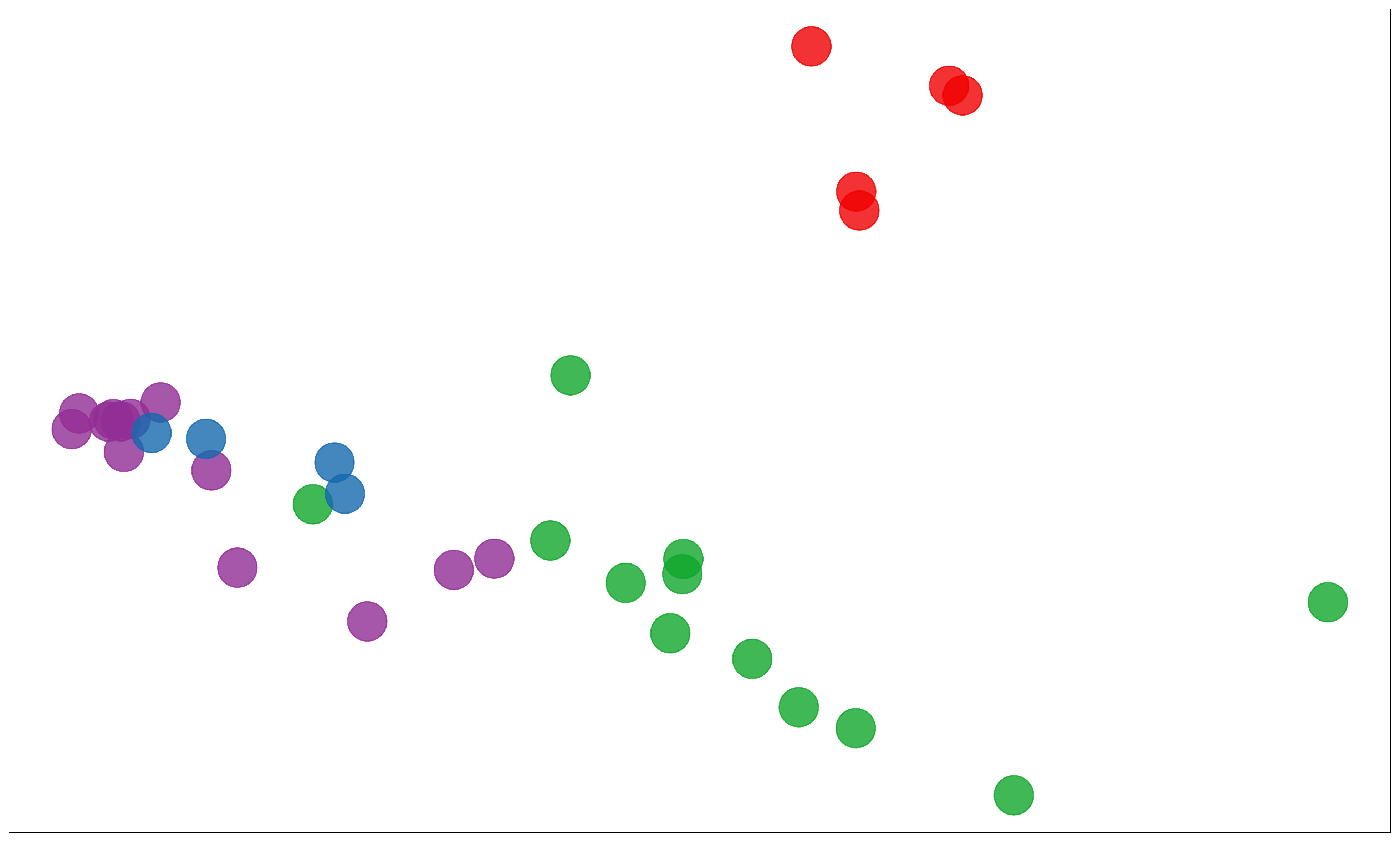}
  \caption{Random-weight GCN.}
  \label{fig:rand_gcn}
\end{subfigure}
\begin{subfigure}[b]{0.230\textwidth}
    \centering
    \captionsetup{size=footnotesize, labelfont=footnotesize}
    \includegraphics[width=\textwidth]{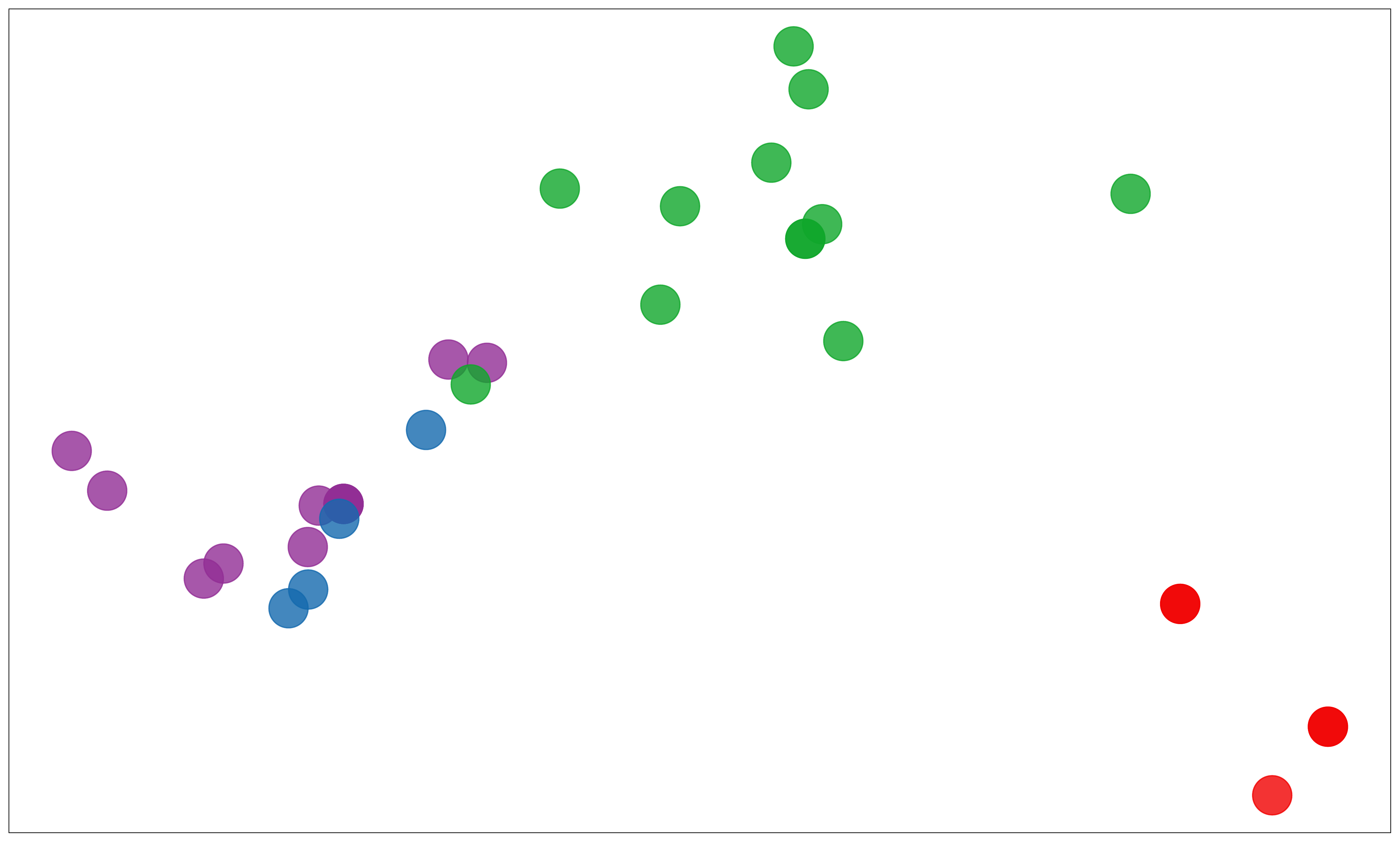}
    \caption{Weight-free GCN.}
    \label{fig:wf_gcn}
\end{subfigure}
\begin{subfigure}[b]{0.230\textwidth}
  \centering
  \captionsetup{size=footnotesize, labelfont=footnotesize}
  \includegraphics[width=\textwidth]{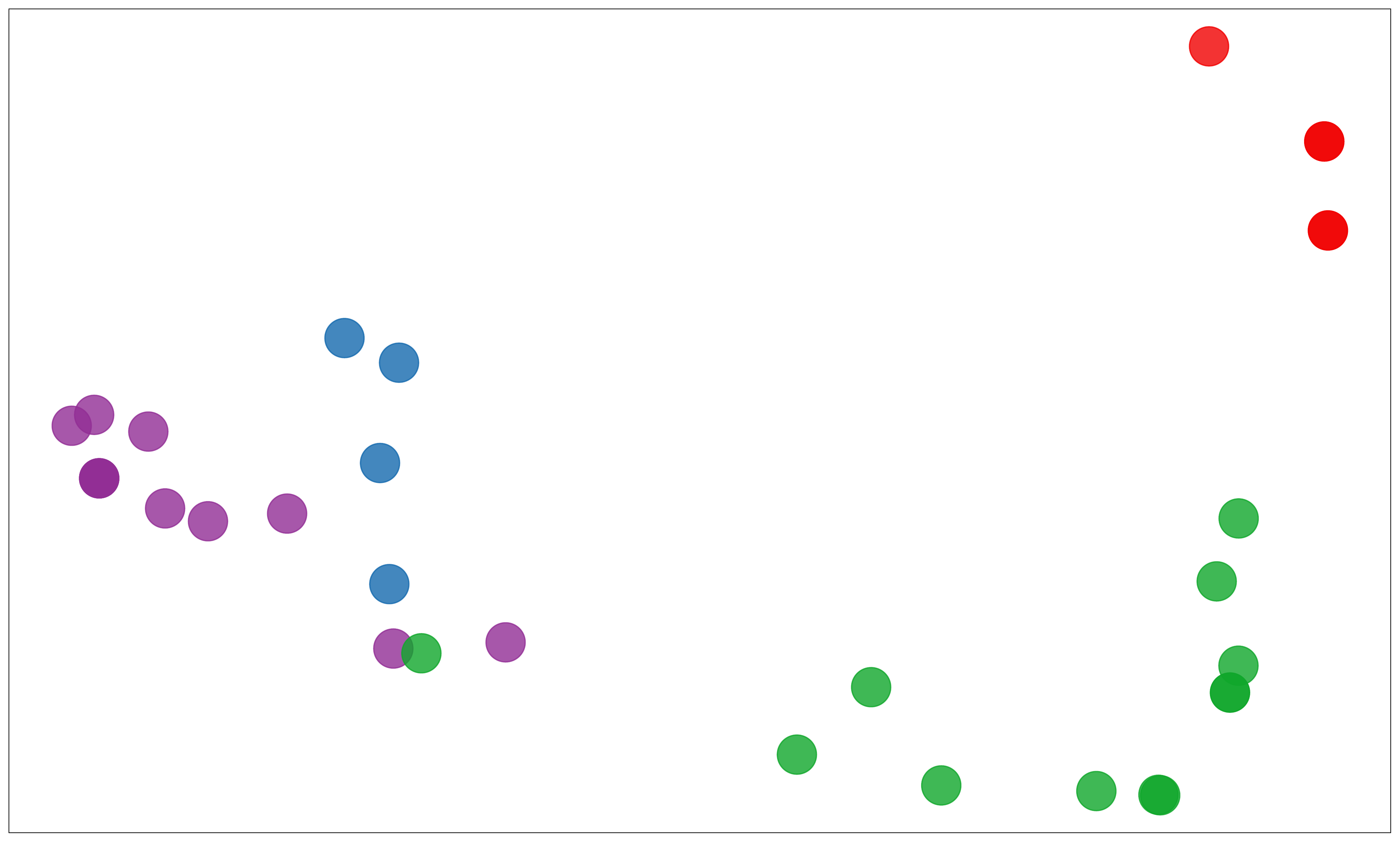}
  \caption{TFGM weight-free GCN.}
  \label{fig:tfgm_gcn}
\end{subfigure}
\caption{Visualization of the Karate club network's node embedding from various training-free GCNs, all have $3$ layers. Nodes of the same color are assigned to the same cluster~\cite{blondel2008fast}. We use PCA for 2-dimensional visualization. Figure~\ref{fig:tfgm_gcn} shows TFGM-flavored weight-free GCN.}
\label{fig:vis_gcn}
\end{figure}

\noindent\textbf{Case Study}. In Figure~\ref{fig:tfgm_vs_basic}, TFGM has shown significant improvement over BasicTFGM. Thus, we conjecture that preserving information from different localities is crucial for obtaining structural representation in the training-free setting. We verify this conjecture in this case study. In Figure~\ref{fig:vis_gcn}, we visualize the node embeddings obtained by training-free GCNs on the Karate club network, a similar case study as in~\cite{kipf2016semi}. Specifically, we present a TFGM-flavored weight-free GCN (Figure~\ref{fig:tfgm_gcn}) to preserve information from different localities. TFGM-flavored weight-free GCN outputs the concatenation of normalized node embeddings from every layer. TFGM-flavored GCN performs better on differentiating blue nodes from purple nodes. In addition, TFGM-flavored GCN and weight-free GCN correctly reflect the symmetry positions of node pairs $(4,10)$ and $(6,5)$ in the network: two pairs of red nodes overlap in Figure~\ref{fig:tfgm_gcn} and Figure~\ref{fig:wf_gcn}.
% Weight-free GCN and random-weight GCN's visualizations are very similar. 

\begin{table}[t!]
\small
\centering
\caption{The accuracy ($\%$) reduction from random-weight GNNs to weight-free GNNs.}
\begin{subtable}[h]{0.32\textwidth}
\caption{DBP15k dataset.}
    \begin{tabular}{lcc}
      \toprule
            & \textbf{GraphSAGE} & \textbf{DGMC}    \\
      \midrule
      ZH-EN & $-0.3$   & $-0.4$ \\
      JA-EN & $-0.2$   & $-0.2$ \\
      FR-EN & $-0.2$   & $-0.1$ \\
      \bottomrule
      \label{tab:dbp15k_reduction}
    \end{tabular}
  \end{subtable}
  \hfill
  \begin{subtable}[h]{0.64\textwidth}
    \caption{PPI dataset.}
    \setlength{\tabcolsep}{1mm}{
  \begin{tabular}{lcccccccccc}
  \toprule
  & \multicolumn{5}{c}{\textbf{Low-conf. Edges}} & \multicolumn{5}{c}{\textbf{Random Rewirement}} \\
  \textbf{Noise Ratio} & \textbf{5\%} & \textbf{10\%} & \textbf{15\%} & \textbf{20\%} & \textbf{25\%} & \textbf{5\%} & \textbf{10\%} & \textbf{15\%} & \textbf{20\%} & \textbf{25\%} \\
  \midrule
  \multicolumn{3}{l}{\textbf{TFGM   pos-enc}} &      &      &      &      &      &      &      &      \\
  GraphSAGE        & $-0.3$        & $-0.3$       & $-0.1$ & $-0.0$  & $-0.0$  & $-0.7$ & $-0.9$ & $-1.1$ & $-1.2$ & $-0.1$ \\
  DGMC             & $-1.0$        & $-2.3$       & $-0.8$ & $-0.7$ & $-0.8$ & $-0.0$  & $-0.7$ & $-1.0$ & $-1.5$ & $-0.9$ \\
  \bottomrule
  \label{tab:ppi_reduction}%
  \end{tabular}
  }%  
  \end{subtable}
  \label{tab:random_reduction}%
\end{table}

\noindent\textbf{Approximation of Random-weight GNNs}. In Section~\ref{sec:rand_vs_free}, we present an interesting proposition that random-weight GNN approximates weight-free GNN for graph matching. To give empirical evidence, we show the performance difference between the weight-free and random-weight versions of the same GNN on DBP15k and PPI datasets (Table~\ref{tab:random_reduction}). We can see that 1) weight-free GNN consistently outperforms random-weight GNN; 2) the real performance reductions ($0\%\sim 2\%$) are negligible compared to the absolute values ($30\%\sim 80\%$). This verifies our claim that random-weight GNN is an unbiased estimator of weight-free GNN.

\subsection{Experimental Setup}
\label{sec:exp_setup}
We implement all the models using PyTorch and PyTorch Geometric on a server with $500$GB memory, two Intel CPU E5-2698 v4 (40 core), and an NVIDIA V100 GPU.

\subsubsection{Baseline GNNs}
\label{sec:baseline_gnns}
We re-implement the GNNs following Algorithm~\ref{alg:tfgm} and remove the nonlinear activations between layers. 
In ablation studies (Section~\ref{sec:ablation}), we demonstrate that the nonlinear activation is not essential for TFGM's performance. We also make necessary changes on architecture of baseline GNNs to adapt to the training-free setting. 

\noindent\textbf{GraphSAGE}~\cite{hamilton2017inductive} usually achieves similar empirical performance with GCN but enjoys efficient deployment via neighbor sampling. Thus, we do not include GCN to avoid redundancy. 
% Although GIN~\cite{xu2018powerful} is the most expressive GNN for graph isomorphism, its guarantee requires the universal approximation property of well-trained MLPs. In the training-free setting, it reduces to GraphSAGE when weights and nonlinearity are removed. 
We employ a weight-free sum aggregator, which is equivalent to the mean aggregator~\cite{hamilton2017inductive} because of the $L2$ normalization in our implementation. We show in Table~\ref{tab:random_reduction} that the weight-free version performs slightly better than the random-weight version.

\noindent\textbf{SplineCNN}~\cite{fey2018splinecnn, fey2020deep} is a graph convolution kernel that can handle spatial geometric relation input. We use it to incorporate the spatial edge feature in natural images. Note that, SplineCNN has a complicated kernel function which is very different from our GNN definition (Section~\ref{sec:problem}). Despite that, we show in experiments that TFGM can significantly improve its performance under the training-free setting. In TFGM, we keep SplineCNN's randomly initialized weights unchanged. We also disable its residual connection. 

\noindent\textbf{DGMC}~\cite{fey2020deep} is the state-of-the-art GNN for graph matching. It iteratively refines the assignments and incorporates the inductive bias of neighborhood consensus~\cite{rocco2018neighbourhood} that can preserve the edge compatibility between two graphs. DGMC uses another GNN as the backbone, which we switch between GraphSAGE or SplineCNN based on the dataset. In TFGM, we replace DGMC's MLP for similarity measurement with training-free cosine similarity.

\subsubsection{Datsets and Hyperparameters}
\noindent\textbf{Supervised Graph Matching}. Keypoint matching is supervised graph matching to find the semantic equivalent keypoints between images of objects. We use the benchmark PascalVOC~\cite{everingham2010pascal} with Berkeley annotation~\cite{bourdev2009poselets}. This dataset is fully supervised with the keypoint annotation for $20$ categories of objects with at most $19$ keypoints. For a fair comparison, our experimental setting is the same as~\cite{fey2020deep}. We use the pre-trained VGG16~\cite{simonyan2014very} to obtain the initial node feature for keypoints. We use the train/test split from~\cite{choy2016universal}. We re-use the source code of ~\cite{fey2020deep} to pre-filter noisy images and keep images with at least one keypoint. We use the anisotropic edge feature due to the better performance. Following~\cite{fey2020deep}, we use $\argmax$ to obtain the approximate solution of the LAP. We use matching accuracy (\%) as the evaluation metric. SplineCNN is only tested on PascalVOC because other datasets have no spatial edge features. 
% \noindent\textbf{Hyper-parameters}. 
For our models, the hidden dimension of GNNs is $512$. Random dimension of DGMC is $128$. GraphSAGE and SplineCNN have $5$ layers. DGMC uses a $1$-layer SplineCNN to perform the same number of refinements ($20$ steps) as in~\cite{fey2020deep}. The $k$NN uses top $10$ neighbors. 

\noindent\textbf{Semi-supervised Graph Matching}. Entity alignment is a semi-supervised graph matching aiming to find equivalent entities between two heterogeneous KGs. We choose a popular benchmark DBP15k~\cite{sun2017cross}, involving three datasets between language pairs: Chinese to English (ZH-EN), Japanese to English (JA-EN), and French to English (FR-EN). Our experimental setup largely follows~\cite{fey2020deep}. We only use the entity names and graph structures in the dataset and do not use the relation types, attributes, and values. The node feature is initialized with the cross-lingual word embedding in~\cite{Xu2019CrosslingualKG}. 
% Following~\cite{fey2020deep}, we implement the fully trained MLP and GraphSAGE with contrastive loss and a heuristic method for negative sampling. 
Following~\cite{fey2020deep, liu2020visual}, we use $\argmax$ to obtain the approximate solution of the LAP. Following~\cite{liu2020visual}, we use Hit@$1$ ($\%$), Hit@$10$ ($\%$), and Mean Reciprocal Rank (MRR) as the evaluation metric.
For our models, the hidden dimension of GNNs is $256$. Random dimension of DGMC is $256$. GraphSAGE has $2$ layers. DGMC uses a $1$-layer GraphSAGE to perform the same number of refinements ($10$ steps) as in~\cite{fey2020deep}.

\noindent\textbf{Unsupervised Graph Matching}. Protein-Protein Interaction (PPI) Network Alignment is unsupervised and aims to find corresponding proteins in networks of different species~\cite{elmsallati2015global, faisal2015post}. Following conventions~\cite{saraph2014magna, vijayan2015magna++}, we use the high-confidence yeast \textit{Saccharomyces cerevisiae} PPI network~\cite{collins2007toward} as the source graph. The target graphs are generated by: 1) adding low-confidence edges into the source graph~\cite{collins2007toward}; 2) randomly rewiring the source graph~\cite{saraph2014magna}. Each target graph has $5$ versions of different noise ratios. For the Low-conf. Edges dataset, we report the average accuracy of $10$ independent runs of TFGM. For the Random Rewirement dataset, we report the average accuracy of TFGM on the $10$ synthetic datasets~\cite{saraph2014magna} at every noise ratio. Following~\cite{zhang2019kergm}, we solve the LAP using the Hungarian algorithm for TFGM and GHOST~\cite{patro2012global}. 
% \noindent\textbf{Hyper-parameters}. 
For our models, the hidden dimension of GNNs is $512$. GraphSAGE has $10$ layers. DGMC uses a $1$-layer GraphSAGE to perform $100$ refinement steps.
% We refer readers to~\cite{bouritsas2020improving} for a detailed discussion of using the counting of substructures as node features. 
We compare with only training-free baselines because there is no supervision.

\begin{algorithm}[t]
  \caption{TFGM (Unsupervised)}
  \label{alg:tfgm}
  \SetAlgoLined
  \KwIn{$\{\mathcal{G}^{(s)}, \mathcal{G}^{(t)}\}$, $\mathrm{GNN}$, $\{W_l\}_{l=1}^{L}$.}
  \KwOut{$S^*$, $U$.}
  \BlankLine
  \For{$\mathcal{G}$ in $\{\mathcal{G}^{(s)}, \mathcal{G}^{(t)}\}$}{
  $(\mathcal{V}, A, X, E)\leftarrow \mathcal{G}$\;
  $H^{(0)}\leftarrow X$\;
  $F^{(0)}_i \leftarrow X_i / ||X_i||_2, \forall i \in \mathcal{V}$\;
  \For{$l=1:L$}{
  $F^{(l)} \leftarrow \mathrm{GNN}(A, H^{(l-1)}, E, W_l)$\;
  $H^{(l)}\leftarrow \sigma(F^{(l)})$\;
  $F^{(l)}_i \leftarrow F^{(l)}_i / ||F^{(l)}_i||_2, \forall i \in \mathcal{V}$\;
  }
  $O \leftarrow [F^{(0)}; F^{(1)}; ...; F^{(L)}]$\;}
 $U_{ij} \leftarrow O^{(s)}_i\cdot O^{(t)}_j\text{, } \forall i\in \mathcal{V}^{(s)}, j\in \mathcal{V}^{(t)}$\;
 $S^* \leftarrow \argmax_{S \in \Tcal} \sum_{i\in\mathcal{V}^{(s)}, j\in\mathcal{V}^{(t)}}S_{ij}U_{ij}$\;
\end{algorithm}

% \begin{minipage}{0.46\textwidth}
\begin{algorithm}[t]
\caption{TFGMws for the supervised setting.}
\SetAlgoLined
\label{alg:supervised_tfgm}
\KwIn{$\{\mathcal{G}^{(s)},\mathcal{G}^{(t)}\}$, $\mathcal{D}=\{\mathcal{G}^{(1)},...,\mathcal{G}^{(N)}\}$, $\mathrm{GNN}$, $\{W_l\}_1^L$.}
\KwOut{$S^*, U$.}
\BlankLine
\For{$\mathcal{G}$ in $\{\mathcal{G}^{(s)}, \mathcal{G}^{(t)}\}$}{
  $(\mathcal{V}, A, X, E)\leftarrow \mathcal{G}$\;
  \For{$\mathcal{G}^{(n)}$ in $\mathcal{D}$}{
    $\hat{S}^{(n)}, U^{(n)} \leftarrow \mathrm{TFGM}(\{\mathcal{G}, \mathcal{G}^{(n)}\}, \mathrm{GNN}, \{W_l\}_1^L)$\; 
  }
  \For{$i$ in $\mathcal{V}$}{
    $\textit{Scores} \leftarrow \{U_{iv}^{(n)}~|~\forall n\in [1,N], v\in \mathcal{V}^{(n)}, \hat{S}^{(n)}_{iv}=1\}$\;
    $\mathcal{L} \leftarrow \{v~|~\forall U_{iv}\in \mathrm{kLargest}(\textit{Scores})\}$\;
    
    $\mathbf{k}_i \leftarrow \sum_{v\in \mathcal{L}}\mathbf{y}_v$ \tcp*{$\mathbf{y}_v$ is the one-hot encoding of node $v$'s label.}
  }
}
$U_{ij}\leftarrow \mathrm{Cos}(\mathbf{k}_i^{(s)}, \mathbf{k}_j^{(t)})\text{, } \forall i\in \mathcal{V}^{(s)}, j\in \mathcal{V}^{(t)}$\;
$S^*\leftarrow \argmax_{S\in \Tcal} \sum_{i\in \mathcal{V}^{(s)}, j\in \mathcal{V}^{(t)}} S_{ij} U_{ij}$\;
\end{algorithm}
% \end{minipage}
% \hfill
% \begin{minipage}{0.46\textwidth}
\begin{algorithm}[t]
\caption{TFGMws for the semi-supervised setting.}
\SetAlgoLined
\label{alg:semi_tfgm}
\KwIn{$\{\mathcal{G}^{(s)},\mathcal{G}^{(t)}\}$, $\mathcal{I}=\{(i,j)|i\in \mathcal{V}^{(s)}, j\in \mathcal{V}^{(t)}\}$, $\mathrm{GNN}$, $\{W_l\}_1^L$.}
\KwOut{$S^*, U$.}
\BlankLine
$(\mathcal{V}^{(s)}, A^{(s)}, X^{(s)}, E^{(s)})\leftarrow \mathcal{G}^{(s)}$\;
$(\mathcal{V}^{(t)}, A^{(t)}, X^{(t)}, E^{(t)})\leftarrow \mathcal{G}^{(t)}$\;
$\hat{X}^{(s)}\leftarrow X^{(s)}$; $\hat{X}^{(s)}_i \leftarrow X^{(t)}_j, \forall (i,j)\in \mathcal{I}$\;
$\hat{X}^{(t)}\leftarrow X^{(t)}$; $\hat{X}^{(t)}_j \leftarrow X^{(s)}_i, \forall (i,j)\in \mathcal{I}$\;
$\hat{\mathcal{G}}^{(s)}\leftarrow (\mathcal{V}^{(s)}, A^{(s)}, \hat{X}^{(s)}, E^{(s)})$\;
$\hat{\mathcal{G}}^{(t)}\leftarrow (\mathcal{V}^{(t)}, A^{(t)}, \hat{X}^{(t)}, E^{(t)})$\;
$\_, U^{(s)} \leftarrow \mathrm{TFGM}(\{\hat{\mathcal{G}}^{(s)}, \mathcal{G}^{(t)}\}, \mathrm{GNN}, \{W_l\}_1^L)$\;
$\_, U^{(t)} \leftarrow \mathrm{TFGM}(\{\mathcal{G}^{(s)}, \hat{\mathcal{G}}^{(t)}\}, \mathrm{GNN}, \{W_l\}_1^L)$\;
$U \leftarrow U^{(s)} + U^{(t)}$\;
$S^*\leftarrow \argmax_{S\in \Tcal} \sum_{i\in \mathcal{V}^{(s)}, j\in \mathcal{V}^{(t)}} S_{ij} U_{ij}$\;
\end{algorithm}
% \end{minipage}

\section{Algorithms}
\label{appendix:alg}
We present the pseudocode of the unsupervised TFGM and the supervised and semi-supervised versions of TFGMws algorithms. Both the supervised and semi-supervised TFGMws rely on the unsupervised TFGM. For all settings, we aim to find the assignment matrix $S^*$ between two input graphs $\mathcal{G}^{(s)}$ and $\mathcal{G}^{(t)}$.

\textbf{Unsupervised TFGM}. Following Algorithm~\ref{alg:tfgm}, we can obtain the assignment matrix $S^*$ by solving the LAP. The similarity matrix $U$ is used in supervised and semi-supervised TFGMws to account for annotation. We use a generalized $\mathrm{GNN}(A, X, E, W)$ which can utilize edge features $E$. The weight $W$ can be discarded if using weight-free GNNs.

First, we normalize each node's feature vector. Then, for each layer GNN, their outputs are (optionally) fed into a nonlinear activation function to be used as inputs in the next layer. After that, we normalize the GNN outputs to ensure equal importance for the comparison of different localities. Finally, we concatenate the GNN outputs from all layers as the final node embeddings. Thus, the dot product in Line $12$ is equivalent to the summation of cosine similarity in Equation~(\ref{eq:cos_gm}). We use an off-the-shelf LAP solver, \textit{e.g.}, Hungarian or $\argmax$, to solve the LAP. We use the same LAP solver as the baselines in experiments for fair comparison.

% We present TFGMws' pseudo-code of utilizing the annotation without training in the supervised and semi-supervised settings. Both the supervised and semi-supervised TFGMws rely on the unsupervised TFGM.

\textbf{Supervised TFGMws}. In keypoint matching of natural images, the training dataset $\mathcal{D}=\{\mathcal{G}^{(1)}, \mathcal{G}^{(2)}, ...\}$ contains graphs of the same object, \textit{e.g.}, motorbike. Algorithm~\ref{alg:supervised_tfgm} present the steps for the supervised TFGMws. The algorithm has two phases: 1) generating $\mathbf{k}_i$ and 2) solving the LAP for $S^*$ with node similarity measured by $\mathrm{Cos}(\mathbf{k}_i^{(s)}, \mathbf{k}_j^{(t)})$, $\forall i \in \mathcal{V}^{(s)}, j \in \mathcal{V}^{(t)}$. To generate $\mathbf{k}_i$ from $\mathcal{G}$, we first conduct graph matching between $\mathcal{G}$ and every graph in $\mathcal{D}$. TFGM returns matrix $U^{(n)}\in \mathbb{R}^{|\mathcal{V}|\times |\mathcal{V}^{(n)}|}$, which measures the similarity scores between nodes in $\mathcal{G}$ and $\mathcal{G}^{(n)}$. Next, for every $\mathcal{G}^{(n)}\in \mathcal{D}$, we select the node that is equivalent to $i$ (based on $\hat{S}^{(n)}$), giving us totally $N$ nodes. Finally, we keep the $k$ nodes with the largest similarity to $i$ and use the sum of the one-hot encoding of these $k$ nodes as $\mathbf{i}$.
% This characteristic of $\mathbf{k}_i$ can be captured by the cosine similarity: $U_{ij}$ will be higher if $i$ and $j$ are the same type of nodes.

\textbf{Semi-supervised TFGMws}. In Algorithm~\ref{alg:semi_tfgm}, we utilize the annotation in semi-supervised graph matching by forcing the known equivalent node pairs to have the same initial feature. Note that, we conduct TFGM twice by comparing $\hat{\mathcal{G}}^{(s)}$ with $\mathcal{G}^{(t)}$ and comparing $\mathcal{G}^{(s)}$ with $\hat{\mathcal{G}}^{(t)}$. Further, we ensemble the two node similarity matrices $U^{(s)}$ and $U^{(t)}$ to achieve better performance.